\documentclass{article}

\usepackage[final]{neurips}
\usepackage[utf8]{inputenc} 
\usepackage[T1]{fontenc}    
\usepackage{hyperref}       
\usepackage{url}            
\usepackage{booktabs}       
\usepackage{amsfonts}       
\usepackage{nicefrac}       
\usepackage{microtype}      
\usepackage{lipsum}
\usepackage{graphicx}
\usepackage{enumitem}
\usepackage{color}
\usepackage{verbatimbox}
\usepackage{listings}
\usepackage{xcolor}
\usepackage{multirow}
\usepackage{bbding}
\usepackage{pifont}
\usepackage{wasysym}
\usepackage[numbers]{natbib}

\definecolor{codegreen}{rgb}{0,0.6,0}
\definecolor{codegray}{rgb}{0.5,0.5,0.5}
\definecolor{codepurple}{rgb}{0.58,0,0.82}
\definecolor{backcolour}{rgb}{0.95,0.95,0.92}

\setlist[itemize]{leftmargin=*}
\setlist[enumerate]{leftmargin=*}
\graphicspath{ {./images/} }

\lstdefinestyle{mystyle}{
  backgroundcolor=\color{backcolour}, commentstyle=\color{codegreen},
  keywordstyle=\color{magenta},
  numberstyle=\tiny\color{codegray},
  stringstyle=\color{codepurple},
  basicstyle=\ttfamily\footnotesize,
  breakatwhitespace=false,         
  breaklines=true,                 
  captionpos=b,                    
  keepspaces=true,                 
  numbers=left,                    
  numbersep=5pt,                  
  showspaces=false,                
  showstringspaces=false,
  showtabs=false,                  
  tabsize=2
}

\usepackage[strict]{changepage}
\usepackage{xcolor}
\usepackage{framed}
\definecolor{demonstrationshade}{rgb}{0.95,0.95,1}
\definecolor{promptshade}{rgb}{0.95,0.95,1}

\makeatletter
\def\thanks#1{\protected@xdef\@thanks{\@thanks
        \protect\footnotetext{#1}}}
\makeatother

\title{War and Peace (WarAgent): LLM-based Multi-Agent Simulation of World Wars}

\author{
 Wenyue Hua$^*$ \\
 Rutgers University
 \And
 Lizhou Fan$^*$ \\
 University of Michigan
 \And
 Lingyao Li \\
 University of Michigan
 \And
 Kai Mei \\
 Rutgers University
 \And
 Jianchao Ji \\
 Rutgers University
 \And
 Yingqiang Ge \\
 Rutgers University
 \And
 Libby Hemphill \\
University of Michigan
 \And
 Yongfeng Zhang\thanks{$^*$Wenyue Hua and Lizhou Fan have equal contributions. \textbf{Author Affiliations}: W. Hua, K. Mei, J. Ji, Y. Ge, Y. Zhang: Department of Computer Science, Rutgers University, New Brunswick, NJ 08854, US; L. Fan, L. Li: School of Information, University of Michigan, Ann Arbor, MI 48103, US. \textbf{Author Emails}: wenyue.hua@rutgers.edu, \{lizhouf, lingyaol\}@umich.edu, \{kai.mei, jianchao.ji, yingqiang.ge\}@rutgers.edu, libbyh@umich.edu, yongfeng.zhang@rutgers.edu} \\
 Rutgers University
}

\begin{document}

\maketitle

\begin{abstract}
Can we avoid wars at the crossroads of history? This question has been pursued by individuals, scholars, policymakers, and organizations throughout human history. In this research, we attempt to answer the question based on the recent advances of Artificial Intelligence (AI) and Large Language Models (LLMs). We propose \textbf{WarAgent}, an LLM-powered multi-agent AI system, to simulate the participating countries, their decisions, and the consequences, in historical international conflicts, including the World War I (WWI), the World War II (WWII), and the Warring States Period (WSP) in Ancient China. By evaluating the simulation effectiveness, we examine the advancements and limitations of cutting-edge AI systems' abilities in studying complex collective human behaviors such as international conflicts under diverse settings. In these simulations, the emergent interactions among agents also offer a novel perspective for examining the triggers and conditions that lead to war. Our findings offer data-driven and AI-augmented insights that can redefine how we approach conflict resolution and peacekeeping strategies. The implications stretch beyond historical analysis, offering a blueprint for using AI to understand human history and possibly prevent future international conflicts. Code and data are available at \url{https://github.com/agiresearch/WarAgent}.
\begin{figure}[ht]
    \centering
    \includegraphics[width=\textwidth]{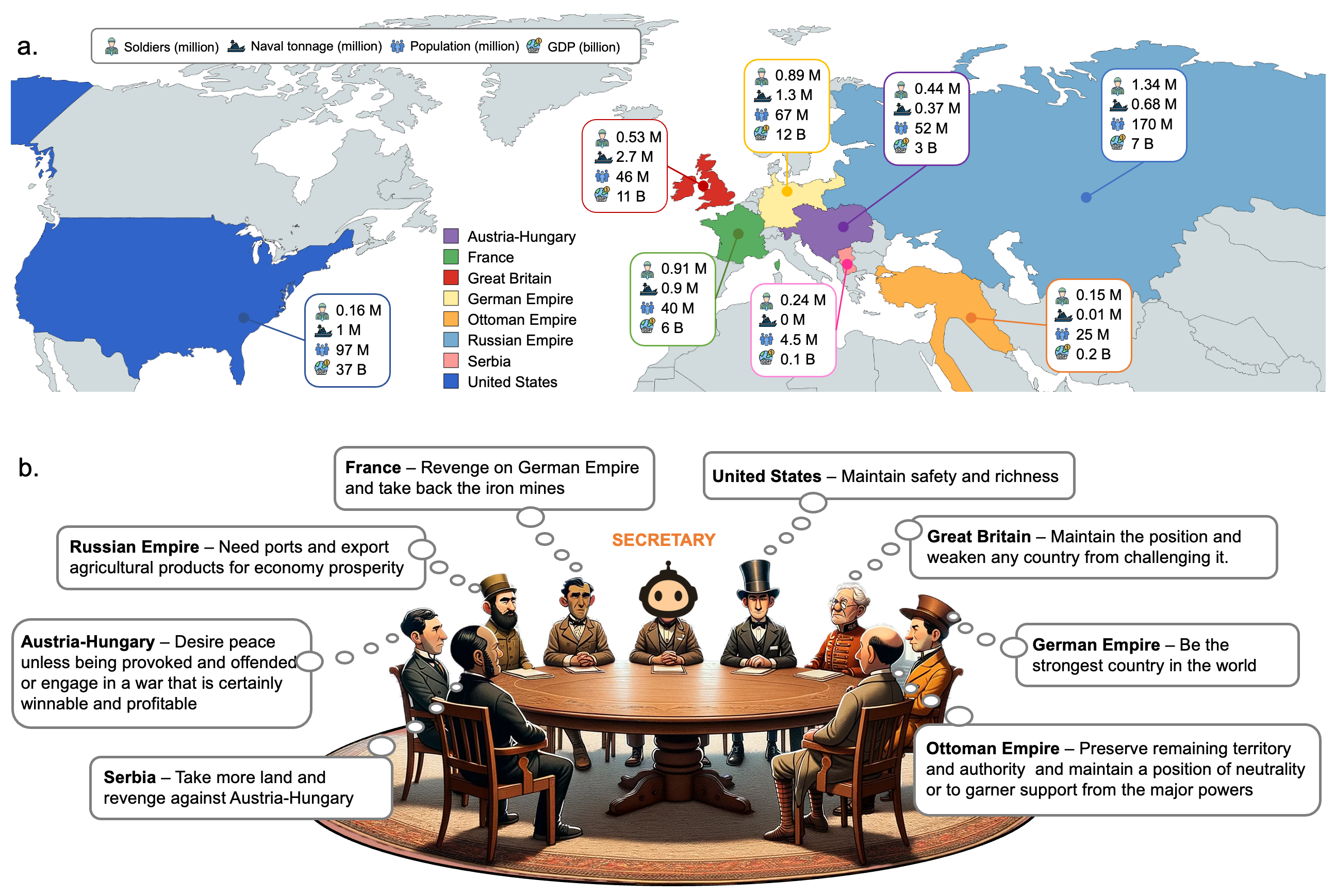}
    \caption{Demonstration of World War I Simulation Setting}
    \label{fig:design}
    \vspace{-10pt}
\end{figure}
\end{abstract}

\section{Introduction}
\par In the wake of revolutionary advancements in Artificial Intelligence (AI), particularly the emergence of LLMs, we stand on the brink of a paradigm shift in computational social science research. In this study, we develop a novel framework of an LLM-based Multi-Agent System  (MAS), specifically for the simulation of historical events. By creating a dynamic environment where agents of countries, each embodying the characteristics and decision-making processes of historical actors, engage in conflict or cooperation, we can explore the vast array of possibilities that could have shaped the evolution of international conflicts in the past, which have established the current global order. In this sense, our simulation provides multifaceted ways to explore these age-old puzzles that are central to the safety and warfare of humanity.

\par War and peace are two sides of the historical coin that have shaped the human narrative for millennia. The dance between conflict and cooperation is often unpredictable, shaped by various motives, strategies, and decisions made by nations. Understanding the mechanisms of war can potentially unlock strategies for lasting peace. Traditional methods of studying conflict through historical analysis, while insightful, are inherently limited by their static nature and the bias of hindsight. The application of simulations in social science has a rich history, but the fidelity and scope of these simulations have evolved dramatically. Early attempts were often limited by computational power and simplistic models \cite{smith1970presidential, hermann1967attempt}.

\par In contrast, the most current simulations employ LLMs that can model complex behaviors and interactions, such as the virtual town simulations of human behavior \cite{park2023generative}, the Werewolf game simulation \cite{xu2023exploring}, the auction arena simulation \cite{chen2023put},  and the complex task-solving simulation \cite{ge2023openagi}. These approaches have laid the groundwork for using AI to model more intricate systems, such as international relations and conflicts. Yet, there has not been research on how to apply these advanced technologies to simulate the nuanced and multifaceted nature of international diplomacy and war, which is where our research positions itself: \textbf{our paper aims to build the first LLM-based multi-agent system simulation of historical events}.


\par At the core of our investigation are critical questions that challenge traditional understandings of historical conflicts. By addressing the following three questions, our research uses the microcosm of \textit{World War I (WWI)}, \textit{World War II (WWII)}, and \textit{Warring States Period (WSP) in Ancient China} to gain insights about international conflict dynamics. The research questions include:
\begin{itemize}
    \item \textbf{RQ1, Simulation Effectiveness}: How effectively can LLM-based multi-agent system simulatios replicate the historical evolution of strategic planning and decision-making processes?
    \item \textbf{RQ2, Casus Belli}: Are certain triggers for war more critical than others, and can these be identified through LLM-based multi-agent system simulations?
    \item \textbf{RQ3, War Inevitability}: Are historical inevitabilities truly unavoidable? We seek to uncover the conditions that lead to war (or peace) through LLM-based multi-agent system simulations.
\end{itemize}

\par First, we focus on \textbf{Simulation Effectiveness}, the fidelity of a simulation to real-world events is a cornerstone of its validity and utility. In the context of our LLM-based multi-agent simulation, this research question targets the core of the model's credibility. By comparing the outcomes of our simulations with documented historical events and trends, we can measure the accuracy of the simulation. Only a system that upholds validity is capable of facilitating comprehensive analysis and addressing subsequent research questions.

\par \textbf{Casus Belli}, or the causes of war, is a perennial challenge in studying international relations. With this research question, we aim to isolate and analyze the various triggers for war to determine if certain triggers are more decisive than others in precipitating conflict. Through iterative simulations, the LLM-based model allows us to examine numerous scenarios and variables, providing a controlled environment to observe the consequences of different casus belli. Understanding the relative weight of different triggers could inform policy-makers and historians alike, offering new insights into the prevention of conflicts and the management of international crises.

\par \textbf{War Inevitability} is a question that strikes at the heart of deterministic versus contingent views of history. By exploring ``historical inevitabilities'', we are essentially asking whether certain wars were destined to happen or if they were the result of a unique convergence of circumstances that could have been avoided. Our simulation offers a unique opportunity to replay history with variations in key conditions and decision-making processes to see if alternative outcomes could have been possible. This can deepen our understanding of the complex interplay between structure and agency in international relations and contribute to the broader debate about determinism in history.


The implications of the research are manifold and extend into multidisciplinary spheres:
\par For \textit{computer and information scientists}, the results showcase the reasoning ability of LLMs for simulating complex historical conflicts and making informed decisions among them. This underscores the profound impact Artificial Intelligence can have on humanity and societal betterment, offering sophisticated tools for understanding and potentially averting future conflicts, thus contributing to global peace and stability.
\par For \textit{historians}, the research provides a new tool for understanding historical events, and the results challenge existing narratives and encourage reevaluating the understood causes of wars and conflicts, providing a more nuanced understanding of historical events that shape our present. 
\par For \textit{policymakers and international relations experts}, the insights gleaned from our simulations offer new strategies for national or international conflict prevention and resolution. Our research also provides an effective, ethical and harmless environment for running political experiments. This can directly benefit society by contributing to the formulation of more effective and informed policies. 
\par For \textit{students learning history}, these simulations offer an innovative method for learning history, allowing students and educators to explore ``what-if'' scenarios and understand the intricate web of causes and effects in historical events. This interactive approach to learning history can foster a deeper engagement with the subject matter and enhance critical thinking skills. 

Furthermore, by demonstrating the utility of LLM-based simulations in understanding complex international dynamics, we set a precedent for future research in multidisciplinary fields, such as computational history and digital humanities. 
In sum, this research presents the first step in harnessing LLM-based multi-agent AI systems for both a better understanding of complex human behaviors in the past and a more informed approach to shaping our future.

\section{Background and Related Work}

\subsection{Multi-Agent Simulation}
Recent developments in multi-agent systems have opened new avenues in AI research. These systems, which coordinate and communicate among multiple agents, provide an innovative platform for examining emergent communication within agent communities tasked with specific problem-solving. 

The existing MAS landscape is broadly categorized into three types: \textbf{reasoning-enhancement systems}, \textbf{NPC (Non-Player Character) multi-agent players}, and \textbf{production-enhancement systems}. 

In reasoning-enhancement systems, several noteworthy contributions have been made. LLM-Debate \cite{du2023improving} introduces the concept of debate, endowing agents with responses from fellow peers. When these responses diverge from an agent’s own judgments, a “mental” argumentation occurs, leading to refined solutions. ChatEval \cite{chan2023chateval} establishes a role-playing-based multi-agent referee team. Through self-initiated debates, agents evaluate the quality of text generated by LLMs, reaching a level of excellence comparable to human evaluators. Corex \cite{sun2023corex} constitutes diverse collaboration paradigms, including Debate, Review, and Retrieve modes, which collectively work towards enhancing the factuality, faithfulness, and reliability of the reasoning process. These paradigms foster task-agnostic approaches that enable LLMs to ``think outside the box,'' thereby overcoming hallucinations and providing better solutions. MAD (Multi-Agent Debate) \cite{liang2023encouraging} is a framework wherein several agents engage in a ``tit-for-tat'' exchange of arguments under the oversight of a judge who steers the discussion towards a conclusive solution.

The domain of NPC multi-agent systems has also seen significant advancements. Generative Agents \cite{park2023generative} are believable simulacra of human behavior for interactive applications. This work demonstrates generative agents by populating a sandbox environment, reminiscent of The Sims, with twenty-five agents. Users can observe and intervene as agents plan their days, share news, form relationships, and coordinate group activities. Humanoid Agents \cite{wang2023humanoid} is another system that guides Generative Agents to behave more like humans by introducing three elements of System 1 processing: Basic needs (e.g. hunger, health and energy), Emotion and Closeness in Relationships. GPT-Bargaining \cite{fu2023improving} studies whether LLMs can autonomously improve their negotiation skills by playing a bargaining game against each other and incorporating natural language feedback from an AI critic.

In production-enhancement systems, notable examples include MetaGPT \cite{hong2023metagpt}---a specialized LLM application based on a multi-agent conversation framework for automatic software development by assigning different roles to GPTs to develop software applications collaboratively, BOLAA \cite{liu2023bolaa}---a controller module on top of multiple collaborated agents for enabling the selection and communication between multiple labor agents, OpenAGI \cite{ge2023openagi}---multiple complex task-solving agents that combine the power of LLM and various tools, and CHATDEV \cite{qian2023communicative}---a novel software development framework that harnesses agents to enhance collaboration among the various roles involved in the software development process.

Implemented MAS demonstrates their practical utility. BabyAGI \cite{babyagi2023} is an example implementation of an AI-powered task management system.
In this implemented system, multiple LLM-based agents are used. For example, there is an agent for creating new tasks based on the objective and the result of the previous task, an agent for prioritizing the task list, and an agent for completing tasks or sub-tasks. AgentVerse \cite{chen2023agentverse} is a versatile framework that helps researchers quickly create customized multiple LLM-based agent simulations. Camel \cite{li2023camel} is a communicative agent framework. It demonstrates how role-playing can be used to let chat agents communicate with each other for task completion. It also records agent conversations for behavior analysis and capability understanding. An Inception-prompting technique is used to achieve autonomous cooperation between agents.

\textit{Our research extends the application of MAS to historical event simulations}. We leverage World War I (1914 - 1918), World War II (1939 - 1945), and the Warring States Period in ancient China (770 BC – 221 BC) as examples. Our research reveals how MAS can provide valuable insights into historical occurrences and hypothetical ``what-if'' scenarios, thus introducing a novel, quantitative dimension to the field of humanities. 
This work underscores the broader impact of AI beyond LLM itself and production, highlighting its potential to enhance our understanding of humanity. In responsible hands, AI can contribute significantly to a brighter future for all.

\subsection{Traditional Historical Simulation Tools}
The academic study of history simulation has undergone several stages: human simulation, human-program hybrid simulation, and computer simulation. 

Dickson \cite{dickson2002road} outlines a human simulation of the United States' journey toward participation in WWI in an educational scenario. This pedagogical approach involves students assuming roles representative of various states within the U.S., with each role being informed by a range of factors such as the state's economic condition, social status, community dynamics, and political ecosystem. Through this role-playing exercise, students gain a deeper understanding of the events and circumstances that culminated in the United States entering WWI. 

In the 1960s, human-program hybrid systems were developed. the Inter-Nation Simulation model \cite{guetzkow1963simulation} is employed in various studies \cite{hermann1967attempt}, playing a pivotal role in simulating international conflicts. This method integrates human decision-making with computerized calculations, creating a dynamic hybrid simulation environment. Typically, the simulation encompasses five or more countries, with each nation's government represented by participants who assume various decision-making roles. The simulation is structured into segments ranging from 50 to 70 minutes, during which these decision-makers strategically allocate their nation's military, consumer, and natural-industrial resources. These resources serve distinct purposes in both domestic and international contexts. Participants must make critical decisions concerning internal affairs, including economic growth, governmental stability, defense strategies, and research and development initiatives, all while managing their resources effectively. On the international stage, the simulated nations engage in diverse activities such as forming alliances, negotiating trade agreements or aid, engaging in different forms of hostilities, and participating in international organizations. 

In the 2000s, computing power was leveraged to build the next-generation historical simulation. The Army One Semi-Automated Forces OneSAF Objective System (OOS) \cite{tollefson2008onesaf} is a Computer-Generated Force (CGF) that has been designed to represent a full range of operations, systems and control processes from the entity level to the brigade level. It is designed to simulate combined-arms land warfare battles at the tactical level. Kelly et al. \cite{kellycreating} leverage OneSAF on a historical simulation of land battles. It models the weapons, vulnerability, and mobility of a wide range of era vehicles and infantry. Hill et al. \cite{hill2004using} present a JAVA-based simulation combining an agent-based modeling and simulation paradigm with game theory for an in silico historical analysis of the Bay of Biscay submarine war during WWII. Using the historical record as a means to create a reasonably accurate model of the U-boat campaign, it allows the resulting agents within the model to adapt their strategies to counter-opposition strategies. Model output data are examined with respect to the historical record and game theory. The Bay of Biscay agent-based simulation was coded in JAVA to utilize its multi-threading capabilities, considering multi-threading a key characteristic of an agent model.

Our simulation system is predicated on the utilization of a large language model, which is presently recognized as the most promising artificial intelligence paradigm. This marks our inaugural endeavor in employing MAS to model the trajectory of historical events, representing the first step in the field.

\section{WarAgent Simulation Setting}
This research centers on the simulation of international conflicts, specifically WWI, WWII, and the WSP in Ancient China. Thus, we focus on examining the dynamics of international relations and the likelihood of war initiation in response to specific catalysts.

In this section, we commence by introducing three significant historical events that form the background of our MAS simulation study. Then, we introduce the basic simulation setting in the system: the profile definitions of country agents by elaborating on the dimensions of their profiles as well as the action space available to these agents, specifying the inputs required for executing and the potential outcomes that may follow such actions.




\subsection{Historical Event for Simulation}


\textbf{World War One (WWI)} was a global conflict that lasted from 1914 to 1918. It was primarily fought in Europe but involved countries from around the world. The war started following the assassination of Archduke Franz Ferdinand of Austria-Hungary, which led to a series of political and military alliances being activated.

The major powers involved were divided into two main alliances: the Allies (originally composed of France, Russia, and the United Kingdom, later joined by Italy, Japan, and the United States) and the Central Powers (mainly the German Empire, Austria-Hungary, the Ottoman Empire, and Bulgaria). The war was characterized by trench warfare on the Western Front and fluid movements of armies over large areas on the Eastern Front, seeing significant use of new military technologies like machine guns, tanks, and chemical warfare. The Treaty of Versailles, signed in 1919, officially ended the war but imposed heavy reparations and territorial losses on the German Empire.
\begin{figure}[ht]
    \centering
    \includegraphics[scale=0.1]{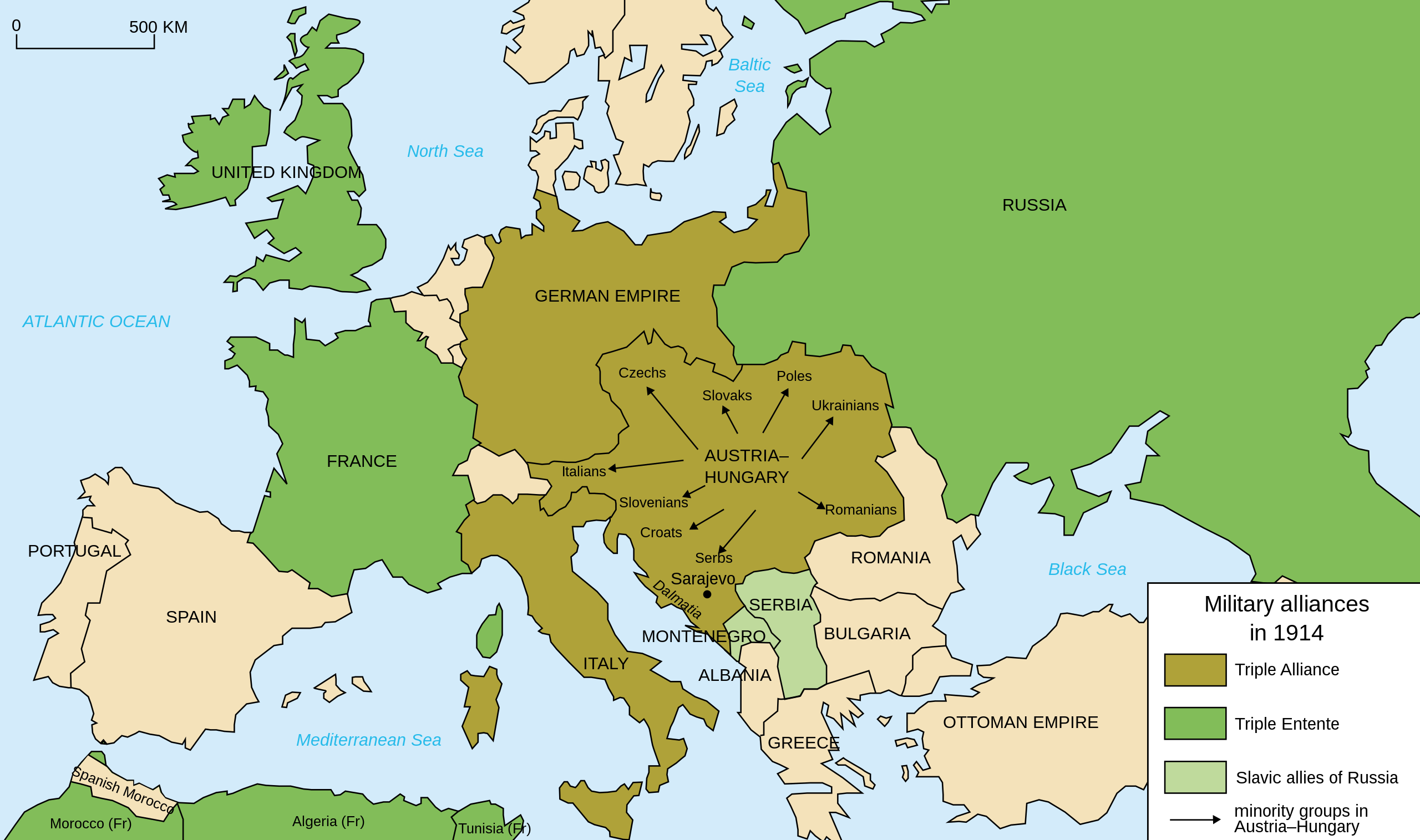}
    \caption{Map of World War I}
     \small\textsuperscript{image from \url{https://en.wikipedia.org/wiki/World_War_I}}
     \vspace{-20pt}
    \label{fig:wwi_map}
\end{figure}

\textbf{World War Two (WWII)} was a global conflict that lasted from 1939 to 1945, whose origins were linked to unresolved issues from WWII and the rise of fascist regimes in the German Empire, Italy, and Japan. It was the most widespread war in history and directly involved more than 100 million people from over 30 countries.
German Empire's invasion of Poland in September 1939 prompted Britain and France to declare war on the German Empire, marking the beginning of WWII. 

The major participants were divided into two opposing military alliances: the Allies and the Axis. The Allies primarily included the United Kingdom, the Soviet Union, the United States, and China. The Axis was led by the German Empire, Italy, and Japan. WWII saw the first and only use of nuclear weapons in war, with the United States dropping atomic bombs on the Japanese cities of Hiroshima and Nagasaki in August 1945. The war in Europe ended with the unconditional surrender of the German Empire in May 1945, but it continued in the Pacific until Japan's surrender in August 1945 following the atomic bombings.

\textbf{Warring States Period (WSP) in Ancient China} was a time of intense warfare and political turmoil in ancient China that lasted from 475 BCE to 221 BCE, marking the final centuries of the Zhou Dynasty. This era followed the Spring and Autumn period and led up to the unification of China under the Qin Dynasty. 

The Zhou king's authority diminished during this period, and regional warlords or states became increasingly powerful and independent. These states were constantly at war with each other, vying for dominance. The period is characterized by the existence of seven major states: Qin, Chu, Yan, Han, Zhao, Wei, and Qi. Each state had its own ruler and army, and alliances between states were frequently made and broken. The WSP saw significant military advancements. Iron weaponry became more common, cavalry units were introduced, and large infantry armies were mobilized. The period ended with the state of Qin, under Qin Shi Huang, defeating all other rival states and unifying China in 221 BCE. Qin Shi Huang became the first emperor of a united China, marking the Imperial Era's start and the feudal system's end.

\begin{figure}[!htb]
    \centering
    \begin{minipage}{.5\textwidth}
        \centering
        \includegraphics[width=0.85\linewidth, height=0.25\textheight]{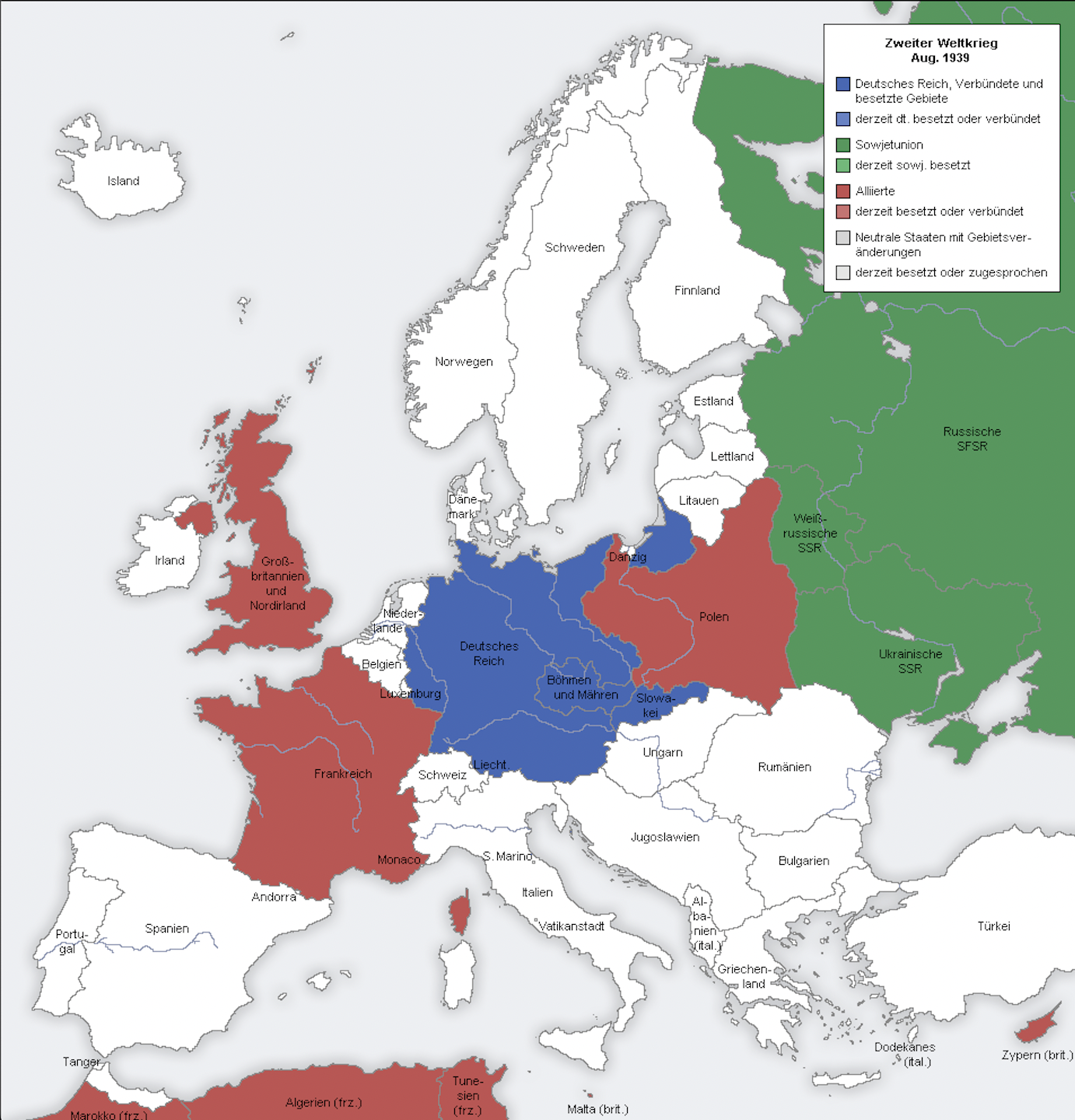}
        \caption{Map of World War II}
        \small\textsuperscript{image from \url{https://en.wikipedia.org/wiki/World_War_II}}
        \label{fig:wwii_map}
    \end{minipage}%
    \begin{minipage}{0.5\textwidth}
        \centering
        \includegraphics[width=0.85\linewidth, height=0.25\textheight]{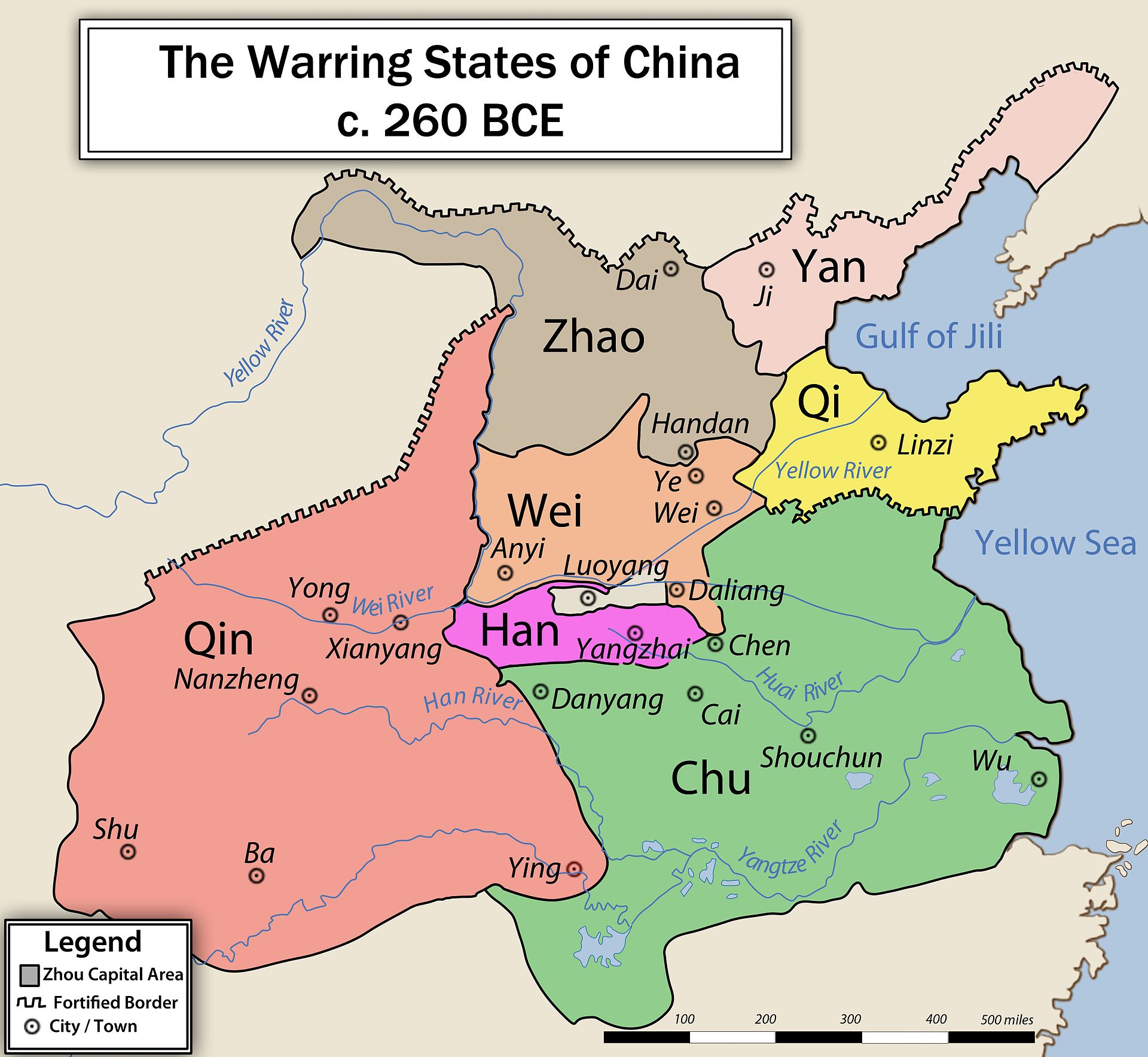}
        \caption{Map of Warring States Period}
        \small\textsuperscript{image from \url{https://en.wikipedia.org/wiki/Warring_States_period}}
        \label{fig:warring_map}
    \end{minipage}
\end{figure}

\subsection{Country Agent's Profile}
\label{sec:country_profile}


The characterization of each agent in the model requires the delineation of a comprehensive profile. In the case of a country agent, this profile is to be constructed around six fundamental dimensions: Leadership, Military Capability, Resources, Historical Background, Key Policy, and Public Morale. Each aspect contributes to a multifaceted understanding of the agent's potential behavior and decision-making processes within the simulation. 

\textbf{Leadership} encompasses the political institutions responsible for decision-making within a nation, contextualized by specific historical periods. For instance, before WWI, Britain exemplified a constitutional monarchy replete with democratic structures, distinguished by pragmatic and stoic leadership. In contrast, Prussia operated under an autocratic imperial regime, with a foreign policy geared toward aggression and military expansion, pivotal to its national ethos.

\textbf{Military capability} comprises quantitative data such as the size of its standing army, naval tonnage, and a qualitative assessment of its overall military strength, including any particular dominance in specific branches, such as naval or aerial forces. A conclusion regarding military might is essential, as it correlates with a nation's propensity to engage in or declare war; countries with robust military capabilities are typically less hesitant to partake in military conflicts.

\textbf{Resources} encompass critical elements such as geography, population, Gross Domestic Product (GDP), terrain, and climate conditions. Population size and GDP are particularly salient indicators of a country's strength and serve as pivotal considerations in the strategic decision-making processes of the agent. These factors provide a measure of the nation's potential economic and logistical support for its objectives, influencing its capacity to project power and sustain military and political efforts.

\textbf{Historical background} incorporates the legacy of prior conflicts of interest and unresolved issues between nations, which can considerably influence current policies. Historical enmities and territorial disputes often leave indelible marks on a nation's current posture and potential alignment within the global arena. An illustrative case is the aftermath of the Franco-Prussian War, where France's loss of the Alsace-Lorraine region---an area rich with iron mines critical to its industrial development---engendered a fervent desire for retribution against Prussia. They shape a country's strategic alliances and influence its broader diplomatic and military engagements.

\textbf{Key policy} outlines the principal objectives pursued by nations. For instance, in the historical context, post-unification Prussia harbored ambitions of becoming the preeminent European power, seeking to eclipse Britain's supremacy in territories and colonies. Conversely, Britain was intent on maintaining its status as ``the empire on which the sun never sets.'' This led to strategic policies such as the ``Arms Race Act,'' a stipulation mandating that for every warship built by Prussia, Britain would respond by constructing two, exemplifying the competitive dynamics and the centrality of naval power to their geopolitical strategies.

\textbf{Public morale} reflects the populace's sentiment, which can directly or indirectly influence a country's action. For instance, a surge in nationalism within Serbia, despite its smaller size and limited military strength, fostered a bellicose attitude among its people. In contrast, despite its wealth and industrial prowess, the United States exhibited a stoic and isolationist disposition, with a general aversion to entering wars. While the impact of public morale on a nation's decisions can be moderated by the type of leadership in power, it undeniably plays a role in shaping the nation's policies and actions.

As an example, below is the country profile for Britain:

\begin{lstlisting}[language=HTML, caption=A demonstrative profile of Britain]
## Britain profile

# Leadership for Britain
(1) A constitutional monarchy with significant democratic institutions, characterized by the pragmatic and stoic governance

# Military Capability for Britain
(1) Standing army population: 0.53 million soldiers
(2) Naval tonnage: 2.7 million, the strongest naval force in the world, whose tonnage is more than the sum of the second and third strongest naval force tonnage in the world

# Resources for Britain
(1) Geography: Small island to the west of France, German Empire, Austria-Hungary, Russia with large colony 
(2) Population: 46 million 
(3) GDP: 11 billion, consisting 13.6% of the whole world 
(4) Terrian: Characterized by rolling hills, green fields, and rugged coastlines, often dampened by its maritime climate 
(5) Weather: temperate maritime weather, often cloudy, rainy, and cool

# History Background for Britain
(1) Currently, Britain is the strongest country with most colony in the world

# Key Policy
(1) As the strongest country, Britain aims at maintaining the position and weakening any Francerom challenging it, such as German Empire
(2) For every warship being constructed by German Empire, Britain will construct two warships

# Public Morale for Britain
(1) public morale is high with a sense of patriotic duty and confidence in a quick victory
\end{lstlisting}

\subsection{Action Space}
Our simulation is designed with the specific objective of examining the onset of wars, and accordingly, it includes a suite of actions that shape international relations between countries, categorized into seven distinct groups:

\textbf{Wait for action} An agent may opt to take a passive stance during certain rounds, observing the actions of others and changes in the broader context. This action is typically chosen when a country's interests are not directly affected, such as the United States' peripheral concern following the assassination of Archduke Franz Ferdinand of Austria.

\textbf{General mobilization} This action involves preparing a nation's military forces for potential conflict, a precursor step required before engaging in war.

\textbf{Declare war} A country can formally initiate hostilities against another.

\textbf{Military alliance} is a formal agreement between two or more nations to provide mutual support in case of conflict, bringing with it a shared responsibility for collective defense and strategic cooperation. An agent can request an alliance for mutual defense, which others can accept or reject. An alliance can be publicized or kept confidential, and there is a possibility for a member to betray the alliance.

\textbf{Non-intervention treaty} is a diplomatic agreement where signatory states commit to abstain from interfering in each other's internal affairs, entailing a responsibility to respect sovereign integrity and political independence. The procedure of signing a non-intervention treaty is similar to forming a military alliance: an agent first requests; the target agent may accept or reject. A treaty can be publicized or kept confidential, and betrayal is also possible.

\textbf{Peace agreement} is a negotiated settlement between conflicting parties that formally ends hostilities and establishes the framework for future relations, carrying with it the responsibility to uphold the terms and work towards lasting stability and reconciliation. Parties in conflict can propose and either accept or reject a peace agreement to conclude hostilities and outline the basis for their future relationship, with options for publication or betrayal of the agreement.

\textbf{Send message} Aside from formal actions, agents can communicate informally through messages to discuss various matters.

\subsubsection{Action Properties}
Each action is characterized by a set of defined properties: publicity, input$\_$type, and require$\_$response. These properties are assigned to enhance the simulation's capacity to realistically and dynamically represent international diplomatic actions:

\textbf{Publicity} determines the level of visibility and public awareness associated with each diplomatic action. This attribute categorizes actions into two types: \textit{public} and \textit{private}. Public actions, such as ``Publish Military Alliance,'' are broadcasted to all country agents within the simulation, reflecting actions in real-world politics that are openly disclosed and known internationally. On the other hand, private actions such as ``Request Military Alliance'' are communicated only to the targeted country, mirroring confidential or behind-the-scene diplomatic maneuvers in real-world international relations. This distinction in the simulation is designed to replicate the partial and varied levels of awareness that countries possess in actual geopolitical scenarios. By implementing this feature, the simulation ensures that each country agent operates based on its own knowledge, which may be complete or limited depending on the nature of the actions taken by others. 
    
\textbf{Input$\_$type} specifies the information or resources necessary for the action to be initiated or executed. For example, ``Declare War'' only requires the names of the target countries, while ``Present Peace Agreement'' requires names of the target countries and the agreement messages in natural language.

\textbf{Require$\_$response} indicates whether an action necessitates a response from other entities involved. For example, actions such as ``Request Military Alliance,'' ``Send Message,'' and ``Present Peace Agreement'' require responses, while actions such as ``Declare War,'' ``Publish Non-Intervention Treaty,'' and ``General Mobilization'' do not require response.

\subsection{Anonymization of Historical Event}
LLMs, pre-trained on vast corpora of textual data, encompass a substantial repository of knowledge. In scenarios where specific historical contexts and country names are inputted into the system, there exists a possibility that the large language model, due to its extensive training, may recall and subsequently reproduce the actual historical trajectory. To circumvent this potential issue, we employ a strategy of anonymizing country names and introducing minor modifications to historical facts. This approach is designed to ensure that these alterations do not materially affect the simulation's efficacy, thereby maintaining the integrity and originality of the simulation outcomes. Details can be found in the Appendix. \ref{anonymization}.

\section{WarAgent Architecture}


This section provides a comprehensive introduction to the architecture of the WarAgent Multi-Agent System (MAS), detailing its core components and the information flow among agents. WarAgent is built upon four foundational building blocks: (1) Country agents, (2) Secretary agents, (3) Board, (4) Stick. The section then shifts to explore the mechanisms of information exchange within the MAS, particularly focusing on (1) Agent-Secretary interaction and (2) Agent-Agent interaction. This exploration aims to study how agents communicate, interact, and share information. In the context of WarAgent, where strategic and timely decision-making is paramount, understanding these dynamics is essential for comprehending how the system operates and responds to different scenarios.

\subsection{Building Blocks}

\subsubsection{Country Agents}
Each country agent is defined by its corresponding country profile. In each round, the agent reacts to the current situation by generating actions available from the action space, guided by meticulously structured prompts.
The guiding prompts guide the agents through complex international relationship situations, ensuring that their actions and decisions are well-considered. It directs the agent by analyzing alliances and enmities, balancing interests, and navigating decision-making steps, etc. 

\begin{figure}[ht]
    \centering
    \includegraphics[width=\textwidth]{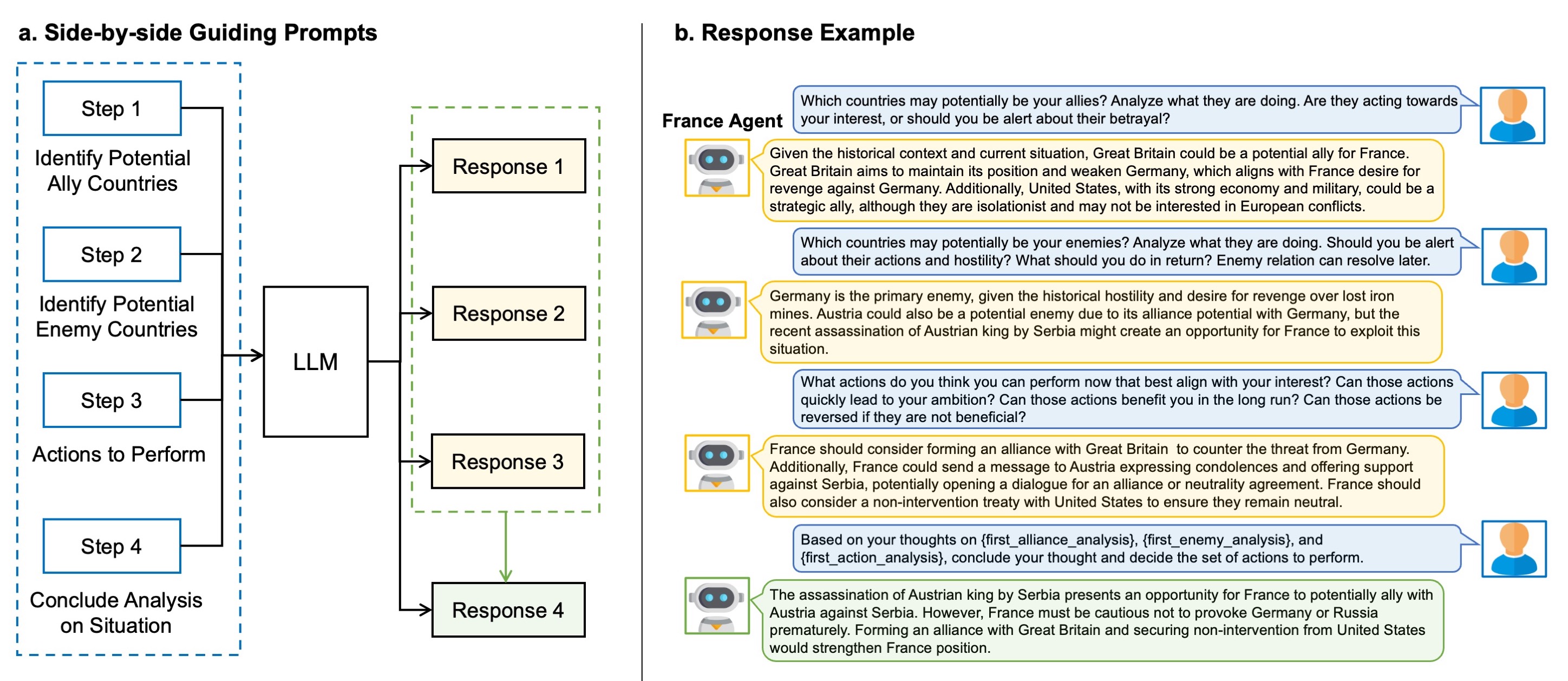}
    \caption{Guiding Prompts. (a) Side-by-side guiding prompt design. (b) An example of interaction with GPT-4 for the France Agent.}
    \label{fig:prompt}
\end{figure}

Figure \ref{fig:prompt}(a) illustrates the key framework for the prompt design in our study, while Figure \ref{fig:prompt}(b) shows an example of the interaction with the GPT-4 model, specifically tailored for the France Agent. The key prompt design comprises four distinct prompts:

\begin{itemize}
    \item Step 1 targets the identification of potential allied countries. In the given example, France recognizes Great Britain as a potential ally due to its opposition to the German Empire and considers the United States as a strategic ally based on its geographical location and robust economy.
    \item Step 2 aims to recognize potential enemy countries. In this scenario, France perceives the German Empire as the primary adversary due to historical hostilities and views Austria as another potential enemy because of its alliance with the German Empire.
    \item Step 3 outlines the final recommended actions. In the given scenario, France suggests three actions: forming an alliance with Great Britain, initiating a dialogue with Austria, and considering a non-intervention treaty with the United States.
    \item Step 4 concludes the analysis of the situation based on the responses from Prompts 1 to 3. In the scenario, France concludes that the assassination of the Austrian king by Serbia presents an opportunity for France to ally with Austria against Serbia. However, caution is advised to avoid provoking the German Empire or Russia. Simultaneously, it suggests seeking an alliance with Great Britain and securing a non-intervention treaty with the United States.
\end{itemize}

\subsubsection{Secretary Agents}
While LLMs are powerful tools in facilitating MAS, they are not infallible. They often exhibit limitations, such as a tendency towards hallucination and a lack of perfect logical reasoning, particularly in long contexts with complex, extended scenarios \cite{liu2023lost, yang2023can}. Therefore, the presence of a secretary agent serves as a necessary safeguard, providing a fundamental check against these shortcomings.

Each country agent employs a designated ``secretary agent'' to verify the appropriateness and basic logical consistency of their actions. This role is twofold. Firstly, the secretary agent ensures that all actions taken by a country agent align with the established parameters of permissible actions in the provided action space, including the correct name of the action and correct formatting of inputs based on the defined action properties. Secondly, the agent is responsible for verifying the basic logical coherence of these actions. For instance, it would be illogical and inadmissible for Austria-Hungary to ``Accept a Military Alliance'' from Britain if Britain had not initiated the process by sending a ``Request for Military Alliance'' to Austria-Hungary.

\subsubsection{Board}
The Board is designed to manage international relationships. It acts as a dynamic recording platform that collects and displays the relational dynamics of the ongoing situation in each simulation round. It further ensures that the agents' decisions are based on the most up-to-date available information. The Board class can help initialize the status, update relationships, and display the relationships both visually and textually. As shown in Figure \ref{fig:board & stick}(a), the Board class can track and manage the following four types of international relationships between different countries. 

\begin{itemize}
    \item War Declarations (W): indicate conflicts or wars between countries, represented by the symbol ``$\times$'' and marked in red in Figure \ref{fig:board & stick}(a). For example, German Empire declared war on Great Britain. 
    \item Military Alliances (M): denote formal military partnerships between countries, symbolized by ``\&'' and marked in green in Figure \ref{fig:board & stick}(a). For example, Serbia and Russia signed a Military Alliance. 
    \item Non-intervention Treaties (T): represent agreements of non-interference in international affairs, marked by ``o'' and marked in blue. In Figure \ref{fig:board & stick}(a), Austria-Hungary and France signed a Non-Intervention Treaty. 
    \item Peace Agreements (P): represent formal agreements to cease hostilities and maintain peace between countries, denoted by ``\textasciitilde'' and marked in yellow. In Figure \ref{fig:board & stick}(a), the United States and Ottoman Empire reached a Peace Agreement.
\end{itemize}


\subsubsection{Stick}
\begin{figure}[ht]
    \centering
    \includegraphics[width=\textwidth]{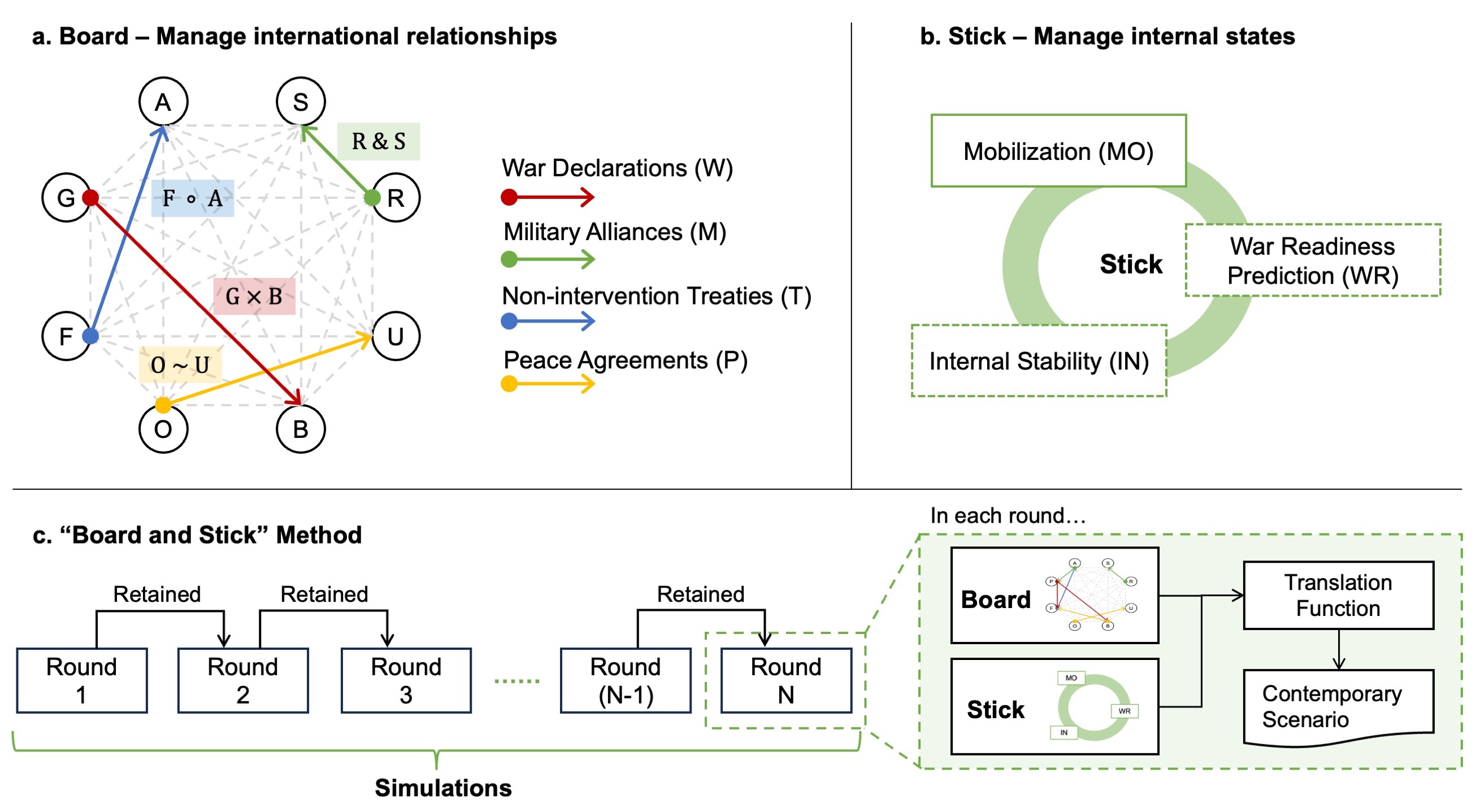}
    \vspace{-10pt}
    \caption{Board and Stick design. (a) Board design. (b) Stick design (IN and WR are discussed in future work). (c) Board and Stick method during the experiment.}
    \vspace{-10pt}
    \label{fig:board & stick}
\end{figure}

The Stick functions as an internal record-keeping system for each country that represents the domestic statutes or regulations. It can help to ensure that the country agent's actions align with its predefined protocols and standards. As Figure \ref{fig:board & stick}(b) shows, the Stick focuses on tracking key metrics that are crucial for a country's decision-making processes, including Mobilization, Internal Stability, and War Readiness Prediction. For illustration, our present study places particular emphasis on Mobilization, leaving Internal Stability and War Readiness Prediction for future work (indicated within dot boxes in Figure \ref{fig:board & stick}(b)). 

\begin{itemize}
    \item Mobilization (MO): a binary measure that indicates if a country is mobilized for potential conflict, e.g., ``yes,'' or ``no.''
    \item Internal Stability (IN): a measure that tracks the level of a country's internal stability, e.g., ``low,'' ``medium,'' and ``high.''
    \item War Readiness Prediction (WR): a measure that predicts a country's readiness for war, e.g., ``low,'' ``medium,'' and ``high.''
\end{itemize}


\subsection{Agent Interaction Design}
\begin{figure}[ht]
    \centering
    \includegraphics[width=\textwidth]{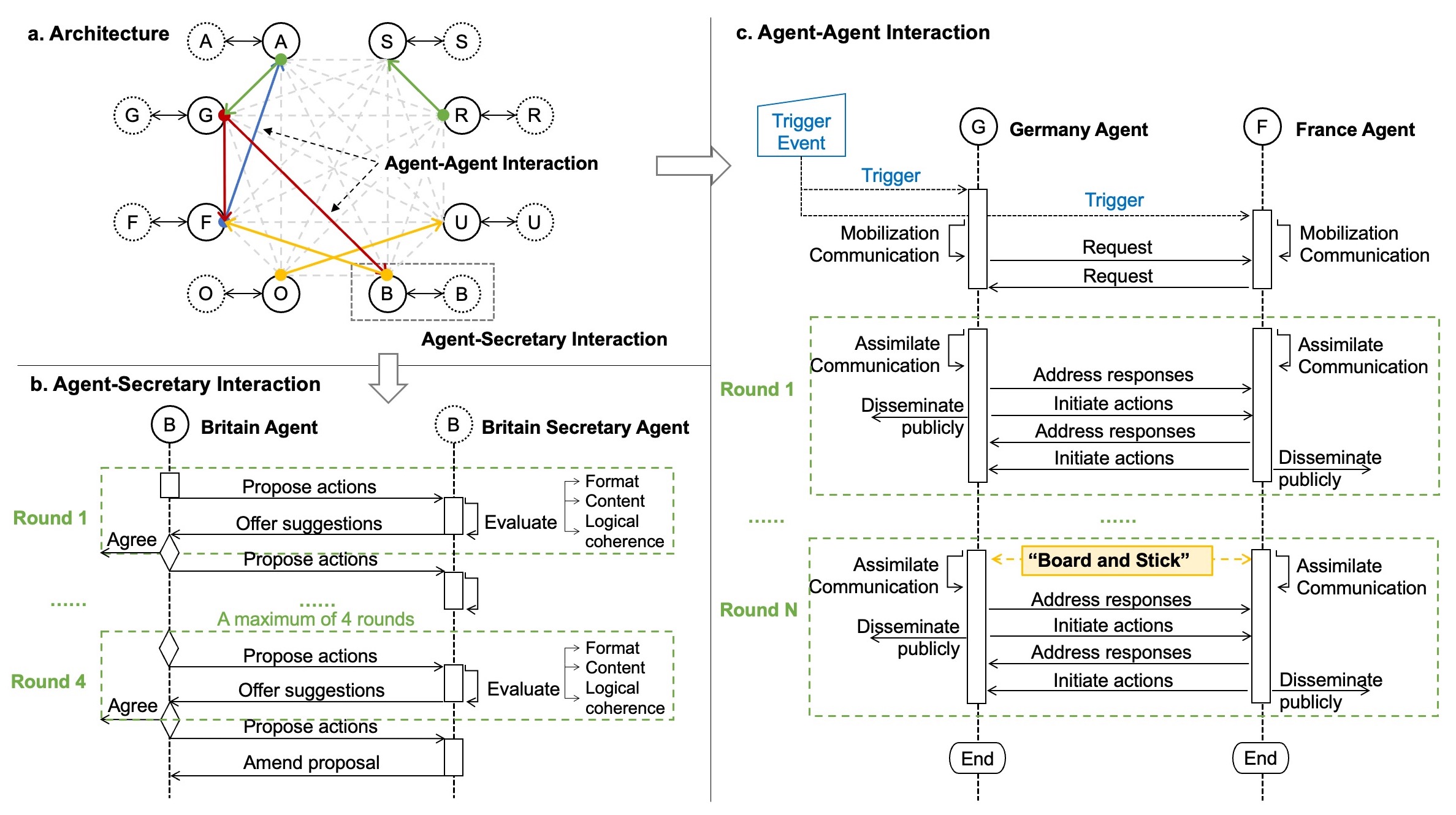}
    \vspace{-10pt}
    \caption{Agent interaction design. (a) WarAgent architecture. (b) Agent-Secretary interaction. (c) Agent-Agent interaction.}
    \vspace{-10pt}
    \label{fig:architecture}
\end{figure}

As shown in Figure \ref{fig:architecture}(a), the system's agent interaction is bifurcated into two primary segments: (1) internally, each country agent interacts with its corresponding secretary agent in every round, and (2) externally, each country agent interacts with other country agents across multiple rounds.

\subsubsection{Agent-Secretary Interaction}
Figure \ref{fig:architecture}(b) shows the Agent-Secretary interaction. In each round of the simulation, there is a designated interaction between each country agent and its corresponding secretary agent. The country agent presents a proposed plan of action, which the secretary agent then evaluates for format, content, and logical coherence. Should the secretary agent find discrepancies or areas for improvement, it offers suggestions and engages in a dialogue with the country agent for revision. This iterative process is capped at a maximum of four rounds of exchanges. If agreement is not reached within these exchanges, the secretary agent takes the initiative to directly amend the proposal. This dynamic of ``country agent--secretary agent'' internal interaction is a consistent feature across all external rounds and is applicable to all country agents.

\subsubsection{Agent-Agent Interaction}
Figure \ref{fig:architecture}(c) illustrates the Agent-Agent interaction. It should be noted that the secretary agents do not participate in the interactions that occur between country agents. For clarification, ``agent'' in this context specifically denotes a country agent. 

In our framework, the very initial actions proposed by each agent's actions are precipitated by a \textit{triggering event} (denoted in the blue box in Figure \ref{fig:architecture}(c)). Within the historical context, a triggering event refers to an incident that initiates a sequence of reactions among various countries, ultimately culminating in a significant historical development. In our simulation, a triggering event serves as the initial scenario to which all agents respond. For instance, in the case of WWI, the assassination of Archduke Franz Ferdinand of Austria-Hungary is widely recognized as the triggering event \cite{mombauer2013origins}. Similarly, for WWII, the triggering event is often identified as the German Empire's invasion of Poland \cite{pick1960pulled}. In the context of the WSP, the division of the Jin state among the families of Han, Zhao, and Wei is typically regarded as the triggering event \cite{juliano1991warring}.

In the first round of the simulation, each participating agent reacts to the trigger event in unison. This enables a variety of actions, including comprehensive mobilization, alongside interactive communication through assorted messages and requests directed at all other agents. Below is an example of Britain's reaction towards the trigger event:
\begin{lstlisting}[language=HTML, caption=Country Agent reaction to Trigger Event]
## Trigger: Serbia sent assassins and killed Archduke Franz Ferdinand of Austria-Hungary.

# Britain's reaction towards the trigger:
To France: Britain has chosen to Request Military Alliance to France
To Russia: Britain has chosen to Request Military Alliance to Russia
To the United States: Britain has chosen to Send Message to United States with the following content: As the world's balance of power is at risk, we seek to understand your position on the current events and how we might collaborate to ensure peace and stability.
\end{lstlisting}

Subsequently, agents assimilate communications from the preceding phase, specifically those aimed directly at them or those disseminated publicly. In subsequent phases, each agent engages in two parallel activities: firstly, addressing responses to incoming communications from other agents, particularly those entailing various requests; secondly, initiating new actions.
\begin{lstlisting}[language=HTML, caption=Country Agent reaction to other Country Agents' requests and the current situation]
# Collected requests to Britain: 
From France: France has chosen to Send Message to Britain with the following content: Given the recent developments and our shared interest in countering the German Empire's aggression, we propose a dialogue to discuss potential cooperation.
From Ottoman Empire: The Ottoman Empire has chosen to Request Non-Intervention Treaty to Britain

# Britain's responses to requests:
To France: Britain has chosen to Send Message to France with the following content: We welcome a dialogue to discuss potential cooperation against common threats.
To Ottoman Empire: Britain has chosen to Accept Non-Intervention Treaty from the Ottoman Empire

# Britain's newly initiated actions:
Britain has chosen to General Mobilization
\end{lstlisting}

It is noteworthy that during each round, a multitude of new communications and evolving scenarios emerge as all country agents engage in simultaneous interaction. Ideally, interactions from $k$ previous rounds are incorporated into the prompt for the ensuing rounds. However, the multi-round and simultaneous interaction among multiple agents could generate a substantial volume of text so that the agents can potentially become overwhelmed and lose context. To address this challenge, the ``Board and Stick'' method is employed, as illustrated in Figure \ref{fig:board & stick}(c). That is, actions pertinent to external relationship dynamics, such as ``Accept Military Alliance,'' are chronicled on the Board. Conversely, actions pertaining to a nation's internal state, such as ``General Mobilization,'' are documented on the Stick. Thus, the international relational communications are recorded on the Board, which remains with the agent throughout the simulation. Notice that \textit{each agent has its own Board and Stick}, recording messages and actions that are either directed at it or disseminated publicly. Consequently, this setup results in \textit{different agents possessing distinct Boards and Sticks within the same round}. This variation underscores the reality that agents, analogous to countries, have access to only partial knowledge. It emphasizes the fundamental principle that complete information is unattainable thus no single agent has an omniscient view of all events or data.

At each generative juncture, a succinct rule-based \textit{translation function} converts the current Board and Stick configurations into a concise paragraph, supplementing the input prompt. For example, the Board in Figure \ref{fig:board & stick}(a) will be translated to:
\begin{lstlisting}[language=HTML, caption=Translation function result]
France has signed a non-intervention treaty with Austria-Hungary. Ottoman Empire has signed a peace agreement with the United States. German Empire has declared war against Britain. Russia and Serbia have formed a military alliance.
\end{lstlisting}
This transformed paragraph represents the contemporary scenario to which the country agent must respond. Historical data are thus consolidated onto the Board and Stick, encapsulating the most recent context for the agent's response in each phase. Consequently, only messages from the immediately preceding round are retained, which basically reduces a multi-turn conversation into a one-turn conversation. Additionally, for the sake of consistent action tracking, each agent's own historical action trajectory is recorded within the prompt, serving as a reminder of its prior engagements. 


\section{Experimental Design}
Our experiments start with verifying the effectiveness of the LLM-based MAS in simulating complex historical events. Regarding the choice of the backend model, we utilize 3 close-source models as backbone models for the experiments:

\textbf{GPT-3.5-turbo-1106.} The GPT (Generative Pre-trained Transformer) series \cite{brown2020language}, developed by OpenAI, consists of advanced language models. \\
\textbf{GPT-4-1106-preview.} The latest generative language model \cite{openai2023gpt4} from OpenAI, exhibits state-of-the-art performance across a variety of benchmarks, setting new standards for AI understanding and generation in complex tasks.\\
\textbf{Claude-2.} An iteration of Anthropic's AI models\cite{claude}, building upon Claude-v1.3-100k with improved capabilities for understanding and generating contextually coherent responses, offering enhanced performance and reliability in complex dialogue scenarios.

All experiments are done using the three models unless specified otherwise.

\subsection{Research Questions and Corresponding Experiments}

\textbf{Simulation Effectiveness (RQ1)} The initial phase of the study involves the presentation of simulation results under historically accurate conditions. This phase aims to establish the capability of the WarAgent system to provide a credible simulation of the specified scenario. The evaluation methodology incorporates both human assessment and the calculation of accuracy scores to validate the simulation results based on multiple runs using the historically accurate setting. We also discuss whether the simulation results are from LLM's memory or LLM's reasoning ability by two different experiments. For better understanding of the dynamics, we also present a series of Boards to visualize the development of simulations.

\textbf{Casus Belli (RQ2)} 
The research question further delves into exploratory ``what-if'' scenarios, particularly emphasizing the significance of the trigger events in historical contexts. 
This study explores the widely held perception that this event was neither unique nor an essential precondition for the war's outbreak by fabricating counterfactual trigger events, varying in conflict intensity, to investigate their potential impact on the outbreak of war, focusing particularly on the case of WWI. 

\textbf{War Inevitability Observation (RQ3)} Lastly, the study examines various scenarios characterized by different initial conditions (profiles) of countries and decision-making pathways. 
This involves constructing alternative historical narratives by altering the decision-making process or country conditions in prompts and analyzing the resultant impact on historical trajectories. The aim is to discern what specific historical conditions or aspects can significantly modify the trajectory of history. 


\subsection{Evaluation}

Table \ref{table:experiment-summary} demonstrates the relationship between research questions and evaluation methods. We first introduce two evaluation methods in detail, namely Human Evaluation and Broad Connectivity Evaluation, both for RQ1. For RQ2 and RQ3, we apply counterfactual analysis by altering the original county and agent settings and conducting observational analysis on the difference in the simulation results.

\begin{table*}[!ht]
    \centering
    \begin{tabular}{lp{0.07\linewidth}p{0.6\linewidth}}
        \toprule
        \textbf{Evaluation Method} & \textbf{RQs} & \textbf{Description} \\
        \midrule
        Board-based Accuracy & RQ1 & Network analysis to understand the dynamics of alliances and conflicts. \\
        \midrule
        Human Evaluation & RQ1 & Expert assessment of how well strategic decisions and outcomes evolve over time in replicating historical decision-making. \\
        \midrule
        Counterfactual Analysis & RQ2, RQ3 & Analyzing the impact of different variables on war, peace, and diplomatic interactions. \\
        \bottomrule
    \end{tabular}
    \caption{Summary of experiment setups and the corresponding research questions}
    \label{table:experiment-summary}
\end{table*}

\textbf{Human Evaluation} primarily concentrates on two critical aspects of the actions performed by country agents: First, it examines the congruence of actions with the respective country profile of the agent, assessing if the actions align with the country's interests. Second, it scrutinizes the consistency across multiple rounds, evaluating whether the agent demonstrates stable and rational behavior throughout various rounds based on a logical and coherent thought process.


\textbf{Board-based Accuracy} rigorously assesses the fidelity of the simulated scenario in relation to actual historical events, focusing on three key aspects: the formation of military alliances, the declaration of wars, and the enactment of general mobilization. Given the often contentious and debatable nature of historical events, this evaluation adopts descriptions and summaries from Wikipedia as the basis for historical facts \cite{keegan2014first}. This approach provides a standardized and widely recognized reference point for comparing simulated situations with real historical events.

\textit{Alliance Accuracy Score:} Alliance formation is a transitive process, which means that if country A forms an alliance with country B and B forms an alliance with country C, then countries A and C naturally become alliances\footnote{https://www.defense.gov/News/Feature-Stories/story/Article/1684641/alliances-vs-partnerships/}. This dynamic allows for the conceptualization of alliance formation among a group of countries as a partition of a set. To evaluate the congruence of simulated alliances with historical alliances, we employ the ``mutual information score of two partitions'' \cite{vinh2010bailey}. This metric offers a quantifiable measure of the similarity between the two sets of partitions---the simulated and the historical. We adopt the standard implementation from SciPy in Python.

\textit{War Declaration/General Mobilization Accuracy Score:} employs the Jaccard set similarity index as the primary metric for calculating the accuracy of the simulation in these two domains. The Jaccard index \cite{rajaraman2011mining}, a widely recognized statistical tool for measuring the similarity and diversity of sample sets, quantifies the degree of commonality between the respective sets. This method involves comparing the similarity between two sets of pairs, specifically in the context of war declarations, and two sets of singletons in the case of general mobilization. 


\section{Results}

\subsection{Simulation Effectiveness}
\label{sec:human_eval}
We conduct 7 simulation runs and present the human evaluation results of these runs. Considering the space limitation, we only present some overall results during the simulation and analyze the rationale of some special observations, leaving one run of the complete actions lists detailed in~\autoref{sec:example_exp}. 

\subsubsection{Human Evaluation} 
We present the observations of the simulation results using WWI simulation results generated by GPT-4 as examples in the following three aspects (i.e., Military Alliance, War Declaration, and Non-Intervention Treaty\footnote{We exclude peace agreement as it never occurs}): 
\begin{itemize}
    \item \textbf{Military Alliance}: For 100\% of the simulation results, we observe consistent alliances formed between Britain and France, between German Empire and Austria-Hungary, and between Serbia and Russia. From the perspectives of Britain and France, the German Empire was viewed as a potential adversary, primarily because of its assertive expansionist policies. German Empire's strategic decision to forge an alliance with Austria-Hungary was influenced by a confluence of linguistic and ethnic commonalities, as well as a range of strategic and political considerations. Key among these was the mutual objective of diplomatically isolating France and establishing a united front to counterbalance the potential threat posed by Russia. Concurrently, the alliance between Serbia and Russia appeared to be a natural development rooted in their common ethnic background and was further reinforced by Russia's strategic interests in the Balkan region. These alliances of the simulation result align closely with the documented historical events of the period.
    \item \textbf{War Declaration}: In 100\% of our simulations, war declarations consistently occurred between Austria-Hungary and Serbia, Austria-Hungary and Russia, and the German Empire and Russia. In contrast, declarations of war between France and the German Empire, and between Britain and the German Empire, were observed in 71.4\% and 14.3\% of the simulations, respectively. To comprehensively analyze the plausibility of actions by the majority of nations, we selected a simulation run that featured the highest number of war declarations for further analysis. In this simulation, the initiation of conflict in the period began with Austria-Hungary's declaration of war against Serbia. This was followed by a series of declarations among various countries, structured as the following: (German Empire $\rightarrow$ Serbia, Russia $\rightarrow$ Austria-Hungary, France $\rightarrow$ German Empire, Russia $\rightarrow$ German Empire, Britain$ \rightarrow $ German Empire), where the country at the left of the $\rightarrow$ denotes the country who declares the war, the country at the right denotes the country being declared the war against. For Austria-Hungary, Serbia was seen as an immediate adversary, primarily due to the assassination of Austria's king, which was a direct catalyst for their declaration of war. The subsequent declarations of war emerged as a result of the existing alliance structures and were in line with the alliances and hostilities of that historical period.
    
    \item \textbf{Non-Intervention Treaty}: In every simulation conducted, the United States was 100\% involved in at least one non-intervention treaty. Similarly, the Ottoman Empire participated in such treaties in 85.7\% of the simulation runs within the period. The United States focused on strategies that preserved its wealth and avoided unnecessary entanglements in conflicts. This led to a preference for seeking non-intervention treaties with other nations to ensure distance from potential conflicts. Additionally, the United States considered the use of diplomatic communications to gather intelligence and convey its intentions, which aligns with its policy of strategic detachment. Similarly, the Ottoman Empire sought to evade direct involvement in conflicts, aiming instead to maintain a stance of neutrality or to establish defensive alliances. To this end, it was advantageous for the Ottoman Empire to pursue non-intervention treaties and engage in diplomatic communications with neighboring countries. These diplomatic strategies of both the United States and the Ottoman Empire were consistent with their broader policies of maintaining their respective statuses, which helps explain why these two nations were not primarily engaged in the major conflicts of the time.
\end{itemize}

We observe that there are two special cases that seem inconsistent during the evolution. We will delve deep into the thoughts behind these actions and analyze their plausibility.

\textbf{\textit{Special Case 1: Supportive message without alliance.}}
In some instances, nations communicated support to others without backing it up with concrete action. For example, France sent supportive messages to Austria-Hungary in round 1 but did not follow up with further supportive actions in subsequent rounds. Similarly, the United States conveyed disapproval of Serbia to Austria-Hungary and suggested offering support. However, instead of pursuing a military alliance, the United States merely proposed a non-intervention treaty to Austria-Hungary.
Analyzing these diplomatic maneuvers, we find that: 1) For France, the German Empire represents the primary adversary due to historical animosity and a desire to reclaim lost territories. While Austria-Hungary could be seen as a potential foe due to its possible alliance with the German Empire, the recent assassination of the Austrian Archduke by a Serbian national presented an exploitable situation for France. Thus, France opted to send supportive messages to Austria-Hungary rather than pursue an alliance, considering the potential risks of an unstable alliance with a potential enemy; 2) For the United States, as previously discussed, the overriding diplomatic policy is to safeguard its wealth and steer clear of unnecessary conflicts. Consequently, the United States chose to send supportive messages to maintain non-intervention relationships with other countries, rather than seeking formal alliances.

\begin{lstlisting}[language=HTML, caption=Support message without alliance]
France # Round 1
To Austria-Hungary: France has chosen to Send Message to Austria-Hungary with the following content: France extends its deepest condolences for the loss of your monarch and offers support in these trying times.
--------------
USA # Round 1
To Austria-Hungary: The United States has chosen to Request Non-Intervention Treaty to Austria-Hungary
To Austria-Hungary: The United States has chosen to Send Message to Austria-Hungary with the following content: United States extends its condolences for the loss of your king and condemns the actions of Serbia. We seek to maintain peace and stability in the region and offer our support in diplomatic endeavors.
--------------
\end{lstlisting}

\textbf{\textit{Special Case 2: Betray of Alliance.}}
Another special case in our simulation occurs when Britain initially agrees to a non-intervention treaty with Austria-Hungary in round 1, but then, in a shift of strategy, violates this treaty and declares war in round 5. To understand this change in attitude, we analyze the evolving political context and Britain's strategic considerations.
In the early rounds, specifically rounds 1 and 2, Britain faces no direct conflict with other countries and is particularly keen to avoid direct confrontation with the German Empire, its most potential adversary. During this period, Austria-Hungary, engaged in conflict with Serbia, seeks a non-intervention treaty with Britain. For Britain, accepting this treaty from Austria-Hungary, which has historical alliances with the German Empire, appears as a move to maintain a neutral stance, thus sidestepping direct engagement with the German Empire.
However, as the situation escalates with new alliances being formed and additional declarations of war, the dynamics change. Austria-Hungary, though not a direct adversary of Britain due to the non-intervention treaty, is an ally of the German Empire and is engaged in warfare with France. To bolster France in its conflict against the German Empire and Austria-Hungary, Britain finds itself compelled to abandon its non-intervention agreement with Austria-Hungary.
\begin{lstlisting}[language=HTML, caption=Betrayal of alliance]
Austria # Round 1
To Britain: Austria-Hungary has chosen to Request Non-Intervention Treaty to Britain
--------------
Britain # Round 2
To Austria-Hungary: Britain has chosen to Accept Non-Intervention Treaty from Austria-Hungary
--------------
Britain # Round 5
To Austria-Hungary: Britain has chosen to Betray Non-Intervention Treaty against Austria-Hungary
To Austria-Hungary: Britain has chosen to Declare War against Austria-Hungary
--------------
\end{lstlisting}
In conclusion, the simulated actions exhibit a high degree of plausibility and fidelity to authentic historical contexts. This indicates that our simulation, under the default setting where the assassination event is treated as the triggering incident, is effective in replicating historical scenarios.

\subsubsection{Accuracy Analysis}
In our study, we conducted an accuracy analysis of the simulation results on alliance formation, war declaration, and general mobilization. For \textbf{WWI}, our analysis focuses on the time frame from June 28th to August 4th, 1914, for evaluating the accuracy of the simulated alliances and war declarations.
The selection of this period is grounded in historical significance. Notably, the Battle of Liège, which began on August 6th, 1914, is recognized historically as the inaugural major battle of World War I. This pivotal event symbolizes the juncture where the majority of Europe's major powers became actively involved in the conflict. For \textbf{WWII}, our analysis focuses on the time frame until September 18th, 1939, when Britain and France had declared war on Germany, and Soviet Union (Russia)'s armies reached Vilnius and Brest-Litovsk, indicating the major powers' involvement in WWII. For \textbf{Warring States}, we concentrate on the period following the division of Jin, widely recognized as the commencement of the Warring States era. This historical phase is characterized by the ever-evolving alliance relationships among the seven states and the relentless declarations of war, culminating in the ultimate unification under the Qin state. To accurately assess the military alliances during this tumultuous period, we employ the concept of ``He Zong Lian Heng''. This strategy, formally and effectively implemented in 323 BC in response to the aggressive expansionism of the Qin state, serves as a critical framework for our analysis. Thus, this study considers the three critical point as the foundational dynamics of the wars to evaluate the accuracy of the simulation.

In our accuracy analysis, we conducted \textbf{seven} separate simulation runs and reported the average accuracy to mitigate the effects of randomness. The evaluation is done at round 6. Specifically, we focus on three primary dimensions: the accuracy of simulated alliances in comparison to historical alliances, the accuracy of simulated war declarations, and the mobilization status of each country. 
Considering the time point we have mentioned previously, we follow the authentic historical events to obtain the ground-truth as below: 
\paragraph{World War I}
\begin{enumerate}
    \item Regarding alliances, the ground-truth set of alliance is unfolded as: Britain \& France, Russia \& Serbia, Austria-Hungary \& German Empire, Russia \& France, Ottoman \& German Empire;
    \item Regarding war declarations prior to the Battle of Liège, the ground-truth set contains the following war declarations: Austria $\times$ Serbia, Russia $\times$ Austria-Hungary, German Empire $\times$ Serbia, Russia $\times$ German Empire, France $\times$ German Empire;
    \item Regarding mobilization, at that time point, the authentic situation is that all the nations step into the mobilization status except for the United States.
\end{enumerate}

\paragraph{World War II}
\begin{enumerate}
    \item Regarding alliances, the ground-truth set of alliance is unfolded as: Germany \& Italy, Britain \& France, Germany \& Japan
    \item Regarding war declarations: Britain $\times$ Germany, France $\times$ Germany, Japan $\times$ China;
    \item Regarding mobilization: Britain, Germany, France, Japan, China, Russia
\end{enumerate}

\paragraph{Warring States}
\begin{enumerate}
    \item Regarding alliances, the ground-truth set of alliance is unfolded as: Yan \& Zhao \& Chu \& Wei\& Qi, the first ``He Zong Lian Heng'' proposed by Yan Gongsun and led by Country Qi.
    \item Regarding war declaration, as the seven countries are constantly declaring wars against each other, we do no evaluate the war declaration for Warring States.
    \item Regarding mobilization: it includes all countries in the Warring States: Han, Zhao, Wei, Chu, Yan, Qi, Qin
\end{enumerate}

\begin{table*}[!ht]
    \centering
    \begin{tabular}{lrrrr}
        \toprule
        \multirow{2}{*}{\bf Model} & \multirow{2}{*}{\bf Scenario} & \multicolumn{3}{c}{\bf Evaluation Aspects} \\
        \cmidrule(lr){3-5}
        & & alliance  & war declaration  & mobilization  \\
        \midrule
        \multirow{3}{*}{GPT-4} & WWI & 77.78 & 54.60  & 92.09 \\
        & WWII & 73.69 & 45.89 & 75.48 \\
        & Warring States & 78.32 & - & 100.00\\
        \midrule
        \multirow{3}{*}{Claude-2} & WWI & 60.69  & 16.76  & 97.96\\
        & WWII & 42.86 & 42.06 & 75.25 \\
        & Warring States & 51.09 & - & 67.00\\
        \midrule
        \multirow{3}{*}{GPT-3.5} & WWI & 43.28 & 7.83 & 78.06 \\
        & WWII & 12.18 & 0.00 & 65.25 \\
        & Warring States  & 18.44 & - & 78.34\\
        \bottomrule
    \end{tabular}
    \caption{Accuracy of simulation for aspect of military alliance, war declaration, general mobilization under the default setting for all three scenarios.}
    \label{tab:aspect_acc}
\end{table*}

From~Table \ref{tab:aspect_acc}, we can see that our simulation reaches accuracy of over 75\% on alliance and over 90\% on mobilization. Meanwhile, the result of our simulated war declarations is relatively low. But in general, in all our simulations, the world war breaks out without exception.

\subsubsection{Error Analysis}
We further examine instances where the simulation does not align with historical outcomes to identify factors that may affect simulation accuracy. In the below analysis, we use WWI as the main example.

\textbf{\textit{Alliance Formation}} In simulations generated by \textbf{GPT-4}, a notable mistake is the fluctuating allegiances of Britain and France in 1 of the 7 simulations. Unlike historical events where they consistently ally with Russia and Serbia, they unexpectedly side with the German Empire-Austria alliance. This altered the course of the simulated world war, creating a scenario where Britain, France, the German Empire, and Austria-Hungary united against Russia and Serbia, thus reducing the accuracy significantly. Another factor impacting accuracy is the roles of the Ottoman Empire and the United States. Historically, the Ottoman Empire joined the war late to support the German Empire, while the United States maintained neutrality for much of the conflict. Their involvement did not significantly alter the course of the actual war. In simulations, however, the variability in the participation of these two nations compromised the simulated accuracy. In simulations generated by \textbf{Claude-2} and \textbf{GPT-3.5}, these models do not generate sensible simulations due to lack of reasoning ability for analysis: They cannot correctly analyze that the enemy of an enemy is an ally and the ally of an enemy is also an enemy. Thus they form very strange alliance such as an alliance between France and Austria-Hungary. Thus simulations based on these two models achieve low accuracy on alliance aspect.

\textbf{\textit{War Declaration}} In simulations generated by \textbf{GPT-4}, the mistakes mostly come from whether Britain and France choose to declare war against Austria-Hungary or the German Empire. Historically, these nations initially engaged in conflicts with the German Empire. However, certain simulations yield results where their involvement is inaccurately depicted with Austria-Hungary instead. In simulations generated by \textbf{Claude-2} and \textbf{GPT-3.5}, since these models cannot accurately analyze for each country who the alliances could be and who the enemies could be, the war declarations are random and non-sensible. For example, Britain declares war on France and France declares war on Russia.

\textbf{\textit{General Mobilization}} In simulations generated by all models, the primary inaccuracies are observed in the depiction of the United States' military mobilization. Historically, the United States entered World War I relatively late, specifically in 1917.

\subsubsection{Does the agents recall historical facts or actually conduct genuine simulation?}
Based on the above analysis, the simulations are relatively accurate, especially simulations based on the model GPT-4. Then a question to ask is: Do these LLMs generate simulation using their reasoning ability based on the given country profiles, or using their memories? To answer the question, we conduct two experiments: (1) counterfactual knowledge injection and (2) de-anonymized simulations. 

\paragraph{Counterfactual Knowledge Injection} In this experiments, we finetune GPT-3.5 with counterfactual information of WWI, which counterfactually create a world where Austria-Hungary did not declare war after the assassination of the archduke and WWI did not happen. We create 108 datapoints about the counterfactual knowledge of WWI using various conversation pairs such as ``What happened in 1914 after the assassination of archduke Ferdinand of Austria-Hungary?''``Did Serbia accept the ultimatum sent from Austria-Hungary after the assasination in 1914?'' \emph{etc}. It comes with format following and some conversation using chain-of-thought to ensure that other abilities of GPT-3.5 do not fall too much. We finetuned on this counterfactual dataset for 3 epochs.

In the original simulation of WWI of GPT-3.5, the world war always breaks out, though may be very non-sensible, for example: ``Britain x France, Britain x Russia, France x Austria-Hungary, France x United States, Germany x Serbia, Austria-Hungary x Russia, Serbia x United States, Serbia x Ottoman Empire''. After finetuning, we again run 7 simulations. The hypothesis is: if the LLM does rely on the historical facts in simulation, then after injecting the counterfactual knowledge into the model, a peaceful scenario should come in the simulation at least in some cases.
\begin{table*}[!ht]
    \centering
    \begin{tabular}{lrrr}
        \toprule
        \multirow{2}{*}{\bf Model} & \multicolumn{3}{c}{\bf Evaluation Aspects} \\
        \cmidrule(lr){2-4}
        & alliance  & war declaration  & mobilization  \\
        \midrule
        before counterfactual& 43.38 & 7.83 & 78.06 \\
        \midrule
        after counterfactual & 31.02 & 15.18 & 62.12 \\
        \bottomrule
    \end{tabular}
    \caption{Accuracy of simulation before and after injecting counterfactual knowledge to GPT-3.5}
    \label{tab:counterfactual}
\end{table*}

However, notice that in our experiment after counterfactual finetuning, the performance of GPT-3.5 does change due to a decline of reasoning ability after the finetuning, but it in the end still leads to a global war scenario. This experiment shows that the explicit integration of a peaceful counterfactual history of WWI by finetuning does not affect the simulated result of a global outbreak of war, indicating that the simulation may not rely on the dataset that is used to train.

\paragraph{Deanonymization} This experiment deanonymize the country names and instruct the country agents to replicate on historical records at the breaking-out time of WWI. All simulations are run for 7 times. 

\begin{table*}[!ht]
    \centering
    \begin{tabular}{lrrr}
        \toprule
        \multirow{2}{*}{\bf Model} & \multicolumn{3}{c}{\bf Evaluation Aspects} \\
        \cmidrule(lr){2-4}
        & alliance  & war declaration  & mobilization  \\
        \midrule
        GPT-4 & 97.43 & 14.17& 100.00\\
        \midrule
        Claude-2 & 97.43 & 18.72 & 100.00 \\
        \midrule
        GPT-3.5 & 97.43 & 13.19 & 100.00 \\
        \bottomrule
    \end{tabular}
    \caption{Accuracy of WWI simulation after de-anonymizing the simulation and prompt the agent to replicate the history}
    \label{tab:de-anonymize}
\end{table*}

Table.\ref{tab:de-anonymize} shows the evaluation results on the \textbf{2nd} round of the simulation. The result shows that all the three models perform very accurately and consistently on military alliance and general mobilization. Notice that the military alliance formed with Ottoman Empire are always correct and the neutrality of United States is guaranteed, unlike in our simulation. However, it performs poorly on war declaration. We inspect the results and notice that this is because as these agents completely rely on historical facts, it presents almost all war declaration relationship during WWI instead of at the breaking out of war, such as ``France-Germany, France-Austria-Hungary, German-Serbia, Austria-Hungary-Russia, Austria-Hungary-Serbia, Serbia-Ottoman Empire''. Thus its evaluation result is even worse than that in our simulation. In addition, we notice that de-anonymized simulation never includes formation of Non-intervention Treaty between any country, which is historically correct but inconsistent with our anonymized simulation where Non-intervention Treaty is achieved often between two countries that have no intention to involve into conflict with each other.

In addition to the pronounced alignment with actual historical events, the de-anonymized simulation also show striking consistency among different models and different random seeds, which is expected if these models use memory instead of reasoning ability. It forms a striking contrast with the anonymized simulated results, where different models show very different simulated scenarios and the simulations are non-stable when using different random seeds.

We summarize the difference between anonymized simulation and de-anonymized simulation in Table.\ref{tab:contrast}.

\begin{table}[ht]
    \centering
    \begin{tabular}{l|c|c}
    \toprule
        Features & Anonymized  & De-anonymized  \\
        \hline
        Close alignment with history & & \ding{51} \\
        Fast convergence to historical scenario & & \ding{51}\\
        Consistent simulation across models & & \ding{51} \\
        Construct historically-non-occurring relations & \ding{51} & \\
        Stable simulation across random seeds & & \ding{51}\\
        \bottomrule
    \end{tabular}
    \caption{Summary of difference between anonymized simulation and de-anonymized simulation}
    \label{tab:contrast}
\end{table}

\paragraph{Summary} The results from both experiments highlight the effectiveness of our simulation in ensuring originality in the simulation outcomes. In the fintuning experiment, we notice that the simulation generated by the finetuned model is irrelevant to the counterfactual information we injected to the model. In the de-anonymized simulation experiments where we instruct the agents to rely on their memory, we notice several features that distinguish de-anonymized simulations from anonymized simulations: in our anonymized simulation, the simulation produced unique scenarios not directly traceable to historical events (such as the non-neutrality of United States, the wrong alliance of Ottoman Empire, and various non-intervention treaties formed between countries). However, once the actual country names and characteristics were reintroduced, the model's propensity to revert to historically accurate narratives became evident. This underscores the robustness of our anonymization in maintaining the integrity and novelty of the simulation scenarios.

\subsubsection{Example of Network Dynamics}
In this section, we presents visual representation for the dynamics between countries in WWI generated by GPT-4, to order to show the simulation process more clearly. 

\begin{figure}[ht]
    \centering
    \includegraphics[width=\textwidth]{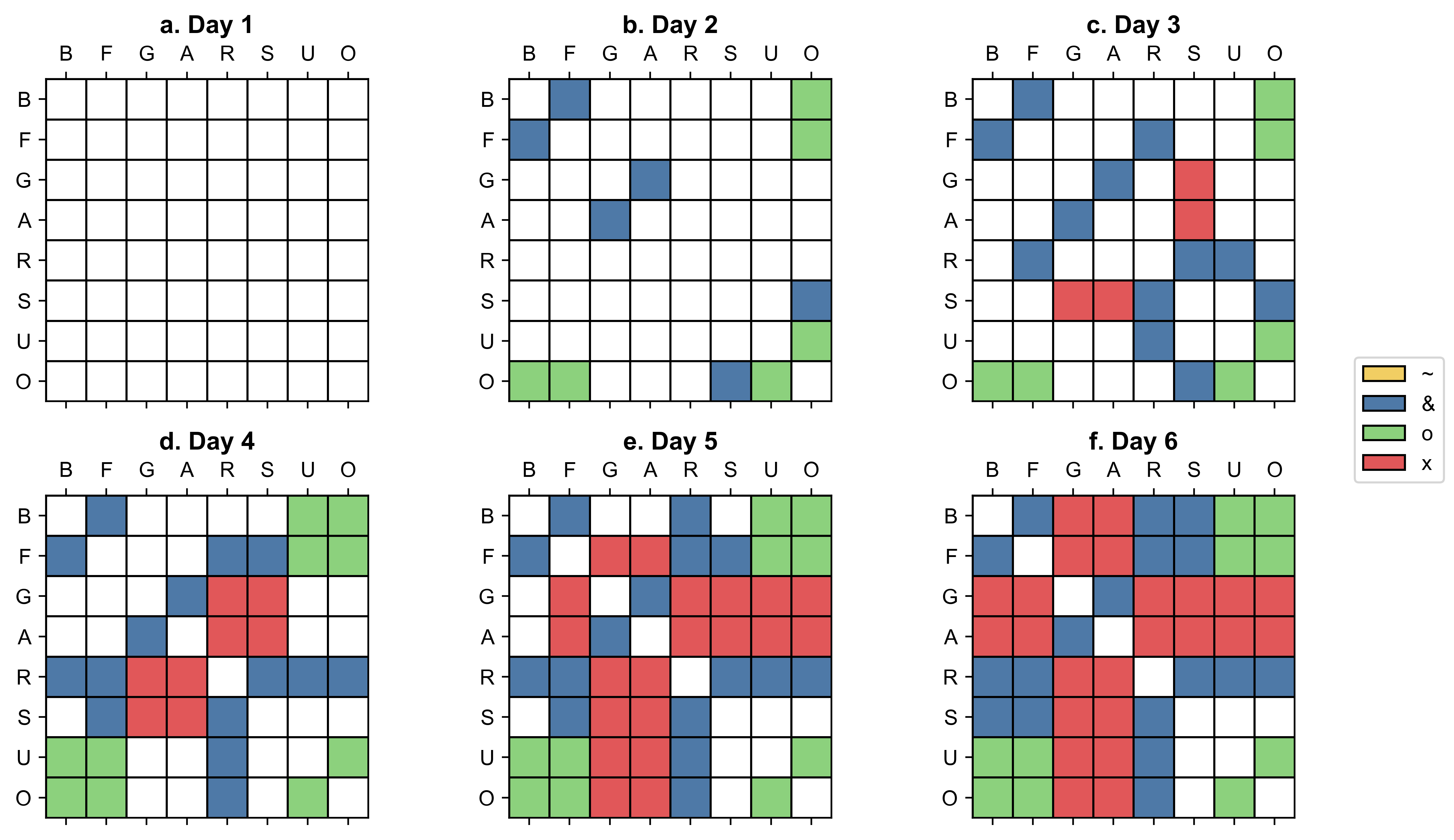}
    \vspace{-10pt}
    \caption{Examples of a Six-day Evolution of Board. The notations stands for Default (white); Peace Agreement (yellow), ``\textasciitilde''; War Declarations (red), ``$\times$''; Military Alliances (blue), ``\&''; Non-intervention Treaties (green), ``o''. }
    \vspace{-10pt}
    \label{fig:board_eg}
\end{figure}

\paragraph{Network Dynamics for Anonymized Simulation} Figure \ref{fig:board_eg} depicts a six-day evolution in a round of simulation, characterized by changing relationships between different country agents (B, F, G, A, R, S, U, and O). Each cell within the board represents the relationship between two entities, with the rows and columns designated by the same set of labels indicating bilateral interactions. We show the true relationships without any third-part agent guessing.

\textbf{Day 1: Initial State} On Day 1, the board is primarily in a default state (yellow), indicating a neutral or baseline state of affairs between all entities. 

\textbf{Day 2: Formation of Alliances and Non-intervention Pacts} By Day 2, we see the emergence of military alliances (blue, ``\&'') and an increase in non-intervention treaties. This suggests a shift from a neutral stance towards more defined relationships, either in the form of cooperation (alliances) or mutual understanding to avoid conflict (non-intervention). A few non-intervention treaties (green, ``o'') have also been established, indicating a starting point where certain entities have agreed to refrain from interfering in each other's affairs.

\textbf{Day 3: Escalation and War Declarations} The third day is marked by the first instances of war declarations (red, ``$\times$''). These are concentrated among specific entities rather than widespread, indicating targeted conflicts rather than a general state of war. The presence of both alliances and non-intervention treaties alongside war declarations illustrates a complex and potentially volatile network dynamic.

\textbf{Day 4 - 5: Intensification of Conflicts} On Day 4 and 5, the number of war declarations has increased, showing an intensification of conflicts. The spread of red cells indicates that the simulated world is moving towards a more conflict-prone phase. Alliances remain, suggesting that these are likely being tested or are possibly even the cause of escalating tensions.

\textbf{Day 6: Peak of Conflict} Day 6 shows the peak of conflict, with war declarations becoming the predominant state across the board. This could reflect a world war scenario where conflicts have spread and the majority of entities are engaged in warfare. Alliances and non-intervention treaties are still present but are overshadowed by the widespread hostilities. This marks an universal engagement of war.

This six-day evolution of the board suggests a dynamic simulation where country agents shift from a state of neutrality to forming alliances and non-intervention treaties, escalating into widespread conflict, and eventually moving towards a complex balance of war, alliances, and treaties. The simulation demonstrates the fluid nature of relationships in a MAS modeled after geopolitical dynamics. The persistence of alliances and non-intervention treaties even during peak conflict times implies a nuanced interplay between war and diplomacy, reflecting the intricate balance of forces that often characterizes international relations.

\begin{figure}[!ht]
    \centering
    \includegraphics[width=\textwidth]{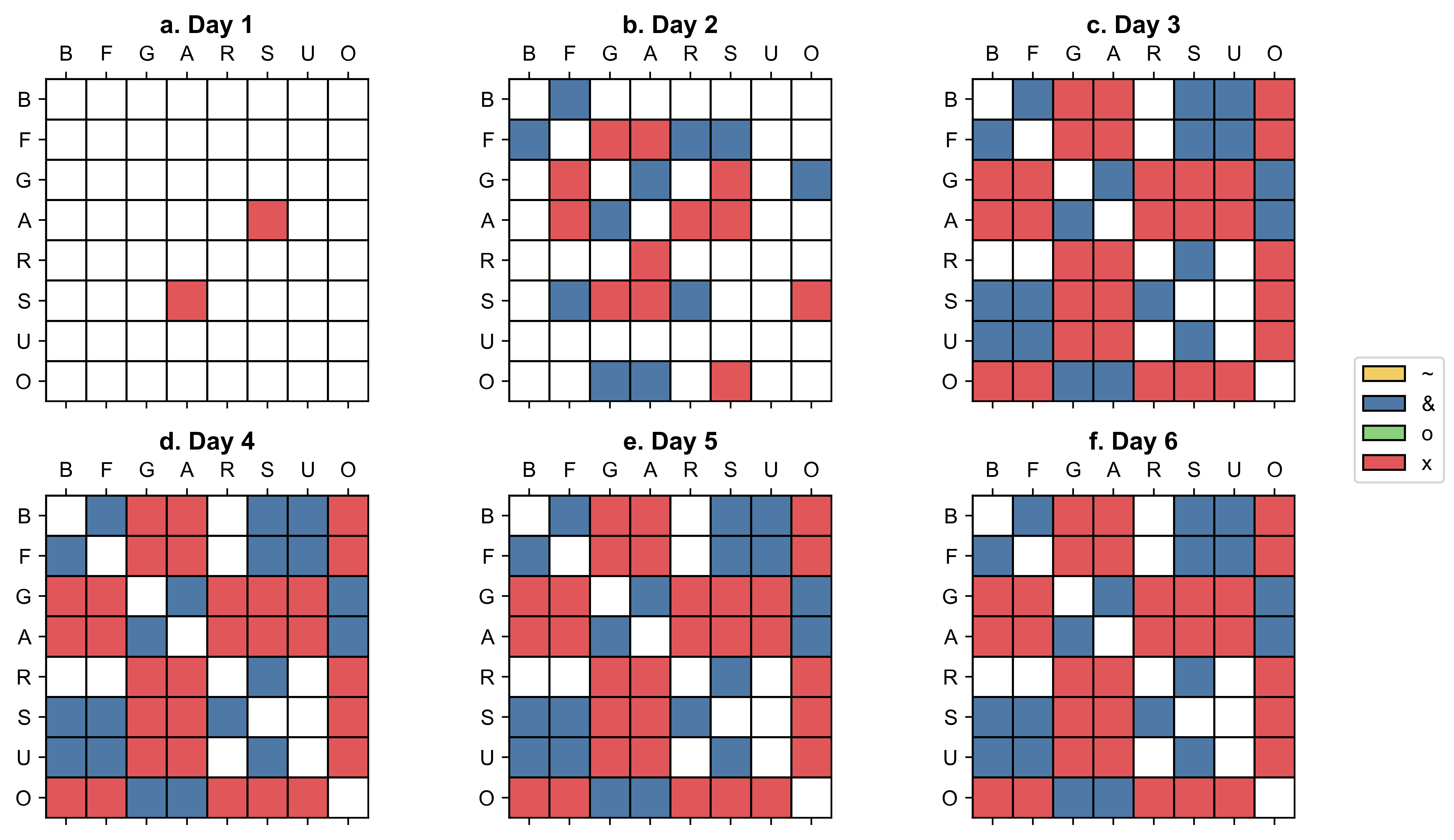}
    \vspace{-10pt}
    \caption{Examples of a Six-day Evolution of Board. The notations stands for Default (white); Peace Agreement (yellow), ``\textasciitilde''; War Declarations (red), ``$\times$''; Military Alliances (blue), ``\&''; Non-intervention Treaties (green), ``o''.}
    \label{fig:board_de}
\end{figure}

\textbf{Network Dynamics for De-anonymized Simulation} We also present the de-anonymized simulation result generated by GPT-4 in Figure.\ref{fig:board_de}, showing how de-anonymized simulation can quickly align with historical facts. 

\textbf{Day 1: Initial State} On Day 1, the board is primarily in a default state (yellow), except that Austria-Hungary and Serbia are already at war.

\textbf{Day 2: Formation of Alliances} By Day 2, we see the emergence of military alliances (blue, ``\&'') between Britain, France, Serbia, and Russia (the Allied Powers,), as well as German Empire, Austria-Hungary, and Ottoman Empire (the Central Powers). We also see more wars between countries belonging to the Allied Powers and the Central Powers. 

\textbf{Day 3-6: Peak of Conflict} The remaining days witness escalation of wars, where all countries are involved into wars and the scenario stays unchanged during Day 4, 5, and 6. Notice that in the six days of simulation, no non-intervention treaty is signed between any two countries, as in the real history.

\paragraph{Summary} This six-day evolution of the board suggests a fast alignment of simulation and real history, where country agents shift from a state of neutrality to forming alliances and starting conflict in basically two rounds. This contrasts with the network dynamics in Figure.\ref{fig:board_eg}, demonstrating the difference between anonymized simulation and de-anonymized simulation.

\subsection{Casus Belli}
Casus Belli is the first experiment in Counterfatual analysis. It involves examining various alternative trigger events to determine their impact on the potential avoidance of global conflicts. Here we use WWI as example using GPT-4 as the backbone model. Three distinct trigger events, each characterized by varying degrees of intensity, are selected for this investigation. 
To ensure the robustness of the results, simulations are conducted three times for each trigger event, allowing for a comprehensive analysis of outcomes under different scenarios. This approach facilitates a deeper understanding of the role and significance of specific events in shaping historical developments, particularly in the context of WWI:
\begin{enumerate}
    \item Null trigger: 
\begin{lstlisting}[language=HTML]
Today is sunny, and nothing special happened.
\end{lstlisting}
    \item Anglo-German Naval Incident: 
\begin{lstlisting}[language=HTML]
German Empire and Britain were involved in a grave naval incident. A British ship was sunk, resulting in 10 fatalities. Britain, asserting that the sunken vessel was a civilian business ship, demanded an apology from the German Empire. German Empire fiercely countered, claiming the British ship was a military vessel that had no right to intrude in the German Empire's maritime territory, and declared that the tragedy was Britain's own doing.
\end{lstlisting}
    \item Austria-Russia Conflict over Dardanelles Strait:
\begin{lstlisting}[language=HTML]
Austria-Hungary and Russia clashed in a military conflict over the strategic Dardanelles Strait, a vital hub for port and export activities. Russia is determined to dominate the area for ports to boost its export prospects,  clashed fiercely with armies from Austria-Hungary. Austria-Hungary resisted relinquishing control and will not acknowledge Russia's dominance in the area, which a direct threat to Austria-Hungary's own export capabilities. Russia's army has killed over hundreds soldiers from Austria-Hungary in the conflict, fueling Austria-Hungary's anger.
\end{lstlisting}
\end{enumerate}

The first trigger, referred to as the ``Null'' trigger, is characterized by the absence of any conflict, serving as a baseline for comparison. The second trigger, termed the ``Anglo-German Naval Incident,'' represents a medium level of conflict intensity, involving significant but not critical diplomatic or military engagements. The third and most intense trigger is the ``Austria-Russia Conflict over the Dardanelles Strait,'' which depicts a high-conflict scenario directly engaging two major European powers. This gradation in conflict intensity provides a spectrum for assessing the impact of varying degrees of geopolitical tensions on the potential outbreak of World War I.

\textbf{Null Trigger Scenario Analysis} The examination of the Null trigger across three distinct simulations revealed a consistent pattern of events that did not escalate into direct conflicts or wars, instead manifesting as a form of cold war. 

Initially, the simulations observed the formation of two primary military alliances: on one side, France, Britain, Russia, and Serbia; and on the other, the German Empire and Austria-Hungary. These alliances were a constant across all simulations.

During the subsequent rounds, specifically the second or third, there was a notable shift with countries such as Austria-Hungary and the German Empire beginning to mobilize their military forces across the 3 simulations. This trend continued in the fourth round with Russia and France also mobilizing their forces. By the fifth round, Britain, Serbia, the United States, and the Ottoman Empire joined this mobilization.

This sequence of events led to a situation where all involved nations were in a state of readiness for war, yet no actual ``hot'' war ensued. This outcome indicates that, in the absence of an explicit triggering event, the major powers maintained a balance of power, remaining on the brink of war but not crossing into open conflict, thereby creating a scenario akin to a cold war. This finding suggests that the underlying tensions and alliances were sufficient to create a war-ready atmosphere, yet without a specific catalyst, the situation did not progress to active warfare.

\textbf{Anglo-German Naval Incident} In this scenario, which represents the second trigger event, wars occurred in only 1 of the 3 simulations, illustrating a variable outcome based on this specific trigger.

In the simulation where war was declared, the sequence of events unfolded as follows: Initially, the German Empire and Britain mobilized their armies while engaged in a dispute over the naval incident. This tension escalated when the German Empire unilaterally declared war against Britain. Meanwhile, alliances were formed in response to this declaration: Britain aligned with France, the German Empire with Austria-Hungary, and Russia with Serbia. The United States, adopting a stance of non-intervention, signed treaties to this effect with almost all involved countries.

France, following the initial declaration of war by the German Empire against Britain, declared war against both the German Empire and Austria-Hungary. Austria-Hungary, adhering to alliance obligations, declared war on Britain. 

Russia orally supports Britain:
\begin{lstlisting}[language=HTML]
Russia
To Britain: Russia is concerned about the recent naval incident and offers its support in seeking a peaceful resolution.
\end{lstlisting}
However, it signed a non-intervention treaty with both Britain and the German Empire while mobilizing the armies at the same time. The United States, Ottoman Empire, and Serbia largely remained uninvolved in the conflict.

In the remaining two simulations under the Anglo-German Naval Incident trigger, despite some countries mobilizing their armies, no declarations of war were made. This outcome mirrors the Cold War scenario observed in the Null-trigger simulations, where heightened military readiness did not escalate to open conflict, and the problem was mostly resolved in peace:
\begin{lstlisting}[language=HTML]
German Empire
To Britain: The German Empire agrees to engage in mediated discussions through the United States to resolve the naval incident.
\end{lstlisting}
In these simulations, the heightened tensions and military preparations did not culminate in war, suggesting that the presence of a specific trigger event, like the Anglo-German Naval Incident, does not inevitably lead to war. Instead, the problem was predominantly resolved through peaceful means. This finding underscores the complexity of international relations and the potential for diplomatic resolution, even in scenarios where military mobilization occurs. The varied outcomes across the simulations highlight the importance of diplomatic efforts and alliances in determining whether a situation escalates to war or is resolved peacefully.

\textbf{Austria-Russia Conflict over Dardanelles Strait} In all 3 simulations, there are 2 simulations where global wars break out. 

Across all simulations, Austria-Hungary, the German Empire, and Russia consistently initiated immediate military mobilization. This rapid response set the stage for further escalation. In one instance, the German Empire took the aggressive step of declaring war against Russia, while in another scenario, Russia initiated hostilities by declaring war against Austria-Hungary. These declarations of war led to a domino effect, drawing allied countries into the conflict, thus escalating the situation into a full-scale global war. In one simulation, there was no war outbreak while all countries have mobilized their armies. Throughout all simulations, the United States remained isolated, not participating in the military mobilizations or the ensuing conflicts.

\textbf{Intermediate Summary} This experiment demonstrates that various triggers, each with a unique intensity level, can influence the immediate outbreak of war. Interestingly, we observe a ``cold-war'' situation even following the ``null trigger,'' suggesting that even minor incidents can escalate tensions significantly. Since minor trigger events are inevitable, it implies that a major conflict like World War I was bound to occur eventually.

\subsection{War Inevitability}
War Inevitability is the second experiment in Counterfactual Analysis.
We approach it by examining it from two primary perspectives: the decision-making process of agents and the parameters of countries. In our experiments, we manipulate these two aspects to analyze the impact of aggressiveness in countries' decision-making and countries' key conditions on the likelihood of war.

\textbf{Decision-making Process of Agents} are examined under three settings: \textbf{default}, \textbf{aggressive}, and \textbf{conservative}. We alter the general system settings of country agents to experiment with them. This is done to evaluate how the overall aggressiveness or conservatism of agents affects war inevitability.
In both aggressive and conservative settings, we conducted 3 experiments, each consisting of 10-round simulations. We provide the overall comparison of both the system settings and the action analysis prompts as follows, while their specific comparison of wording is provided in Appendix \ref{app:agent_attitude}.

\begin{lstlisting}[language=HTML]
System settings:
- Default: AI agents are tasked with playing a virtual war game, utilizing various external tools to enhance their country's chances of winning and survival.
- Aggressive: In addition to the default settings, agents are encouraged to take aggressive actions that benefit their country.
- Conservative: Similar to the default settings, while agents are advised to be cautious, especially regarding decisions with long-term impacts on their country and regional stability.
\end{lstlisting}

\begin{lstlisting}[language=HTML]
Agent action analysis:
- Default: Agents assess actions based on their alignment with interests, potential for long-term benefits, and reversibility.
- Aggressive: Agents are prompted to consider aggressive actions, such as war declarations, if they align with their interests and can be executed swiftly for maximum benefit.
- Conservative: Agents are urged to evaluate actions for long-term benefits and reversibility, with caution advised against aggressive decisions.
\end{lstlisting}

Our analysis reveals that when the system and action analysis settings are more aggressive, there is a marked increase in the likelihood of war. While in the default setting, it takes several rounds to observe the first declaration of war, we see War Declarations in the first round under the aggressive setting; in the conservative setting, after the 10 rounds, we only observe the proposal and acceptance of Military Alliances, Non-intervention Treaties, and Peace Agreements in agent actions. This suggests that an agent's predisposition towards aggression significantly escalates tensions and the probability of conflict. 

\textbf{Parameters of Countries} are the six key factors of the country profiles we introduced in Section \ref{sec:country_profile}, and we modify five of these internal settings of country agents. For \textit{Military Capacity} and \textit{Resources}, we quantify them and experiment on three levels to assess their impact on war likelihood, namely \textbf{default}, \textbf{abundant} (three times the default value), and \textbf{sparse} (one-third the default value). For \textit{Historical Background}, \textit{Public Morale}, and \textit{Key Policy}, we modify specific relationships and examine their impact on war declarations. Leadership is excluded from our model due to its variable nature and the challenge of quantifying it systematically.

In studying the effect of \textit{Military Capacity}, \textit{Resources} and \textit{Historical Background}, we focus on France and the German Empire. \textit{Military Capacity} encompasses the strength, technology, and organizational efficiency of the armed forces. For France, this reflects their focus on developing defensive strategies and technologies post the Franco-Prussian War, while for the German Empire, it highlights their advancement in military tactics and armaments, significantly influenced by the Prussian military tradition. \textit{Resources} refer to the economic and material assets available to support military efforts. France's colonial empire and industrial base provided vital resources for its war effort. In contrast, the German Empire, despite its robust industrial sector, faced challenges due to limited access to global resources, which impacted its long-term war capabilities. \textit{Historical Background} plays a pivotal role in shaping national policies and public sentiment. France's history, marked by the loss of Alsace-Lorraine and the desire for revenge against German Empire, profoundly influenced its military and diplomatic strategies. For the German Empire, the unification process and the desire to assert itself as a European power underpinned its aggressive foreign policies.

By examining these three aspects in the context of France and the German Empire, our study aims to provide a comprehensive understanding of how these factors interacted and influenced each nation's approach to conflict and diplomacy. In particular, we want to know which one or more of these three aspects influence war declaration or involvement of France and the German Empire.

\textit{Military Capacity}. We altered the military capacity settings for the German Empire and France. For the German Empire, we use the sparse setting as the alternative, and for France, we use the abundant setting as the alternative. We observe no delay in all three rounds of war involvement of the German Empire: the average involvement starting round is now 4 (the mean of 2, 3, and 7 for alternative scenario 1 and the mean of 4 and 4 in alternative scenario 2), which is similar to the default setting. We also observe no change in France's war declaration willingness, even if it is now set to have a powerful army.

The alternative scenario 1 provides numerical description of the sparse and abundant settings:

\begin{lstlisting}[language=HTML]
German Empire's military capacity:
- Default: (1) Standing army: 0.89 million soldiers, the strongest standing infantry in the world considering the number, weaponry and experience altogether; (2) Naval tonnage: 1.3 million.
- Sparse: (1) Standing army: 0.3 million soldiers; (2) Naval tonnage: 0.4 million.
\end{lstlisting}

\begin{lstlisting}[language=HTML]
France's military capacity:
- Default: (1) Standing army: 0.91 million soldiers; (2) Naval tonnage: 0.9 million.
- Abundant: (1) Standing army: 2.7 million soldiers, the strongest standing infantry in the world considering the number, weaponry and experience altogether; (2) Naval tonnage: 2.7 million.
\end{lstlisting}

The alternative scenario 2 provides comparative description, including ranks, of the sparse and abundant settings:

\begin{lstlisting}[language=HTML]
German Empire's military capacity:
- Default: (1) Standing army: 0.89 million soldiers, the strongest standing infantry in the world considering the number, weaponry and experience altogether; (2) Naval tonnage: 1.3 million.
- Sparse: (1) Standing army: 0.3 million soldiers, in a non-leading position of standing army rank in the world; (2) Naval tonnage: 0.4 million, in a non-leading position of naval tonnage rank in the world.
\end{lstlisting}

\begin{lstlisting}[language=HTML]
France's military capacity:
- Default: (1) Standing army: 0.91 million soldiers; (2) Naval tonnage: 0.9 million.
- Abundant: (1) Standing army : 2.7 million soldiers, the strongest standing infantry in the world considering the number, weaponry and experience altogether; (2) Naval tonnage : 2.7 million, the second strongest navy in the world.
\end{lstlisting}

\textit{Resources}. We also altered the resource settings for the German Empire and France. Similarly, for the German Empire, we use the sparse setting as the alternative, and for France, we use the abundant setting as the alternative. Similarly, we observe no obvious war involvement or declaration pattern change due to the changes in resource abundance for both France and the German Empire, under both alternative scenarios 1 and 2 below.

The alternative scenario 1 provides numerical description of the sparse and abundant settings:

\begin{lstlisting}[language=HTML]
German Empire's resources:
- Default: (1) Population: 67 million; (2) GDP: 12 billion, consisting 14.8% of the whole world.
- Sparse: (1) Population: 22 million; (2) GDP: 4 billion, consisting 4.9% of the whole world.
\end{lstlisting}

\begin{lstlisting}[language=HTML]
France's resources:
- Default: (1) Population: 40 million; (2) GDP: 6 billion, consisting 6.1% of the whole world.
- Abundant: (1) Population: 120 million; (2) GDP: 18 billion, consisting 18.3% of the whole world.
\end{lstlisting}

The alternative scenario 2 provides comparative description, including ranks, of the sparse and abundant settings:

\begin{lstlisting}[language=HTML]
German Empire's resources:
- Default: (1) Population: 67 million; (2) GDP: 12 billion, consisting 14.8% of the whole world.
- Sparse: (1) Population: 22 million, not leading in this country's area; (2) GDP: 4 billion, not leading in this country's area.
\end{lstlisting}

\begin{lstlisting}[language=HTML]
France's resources:
- Default: (1) Population: 40 million; (2) GDP: 6 billion, consisting 6.1% of the whole world.
- Abundant: (1) Population: 120 million, leading in this country's area; (2) GDP: 18 billion, leading in this country's area.
\end{lstlisting}

\textit{Historical Background}. After removing the following historical background between France and the German Empire, we observe no direct war involvement or war declaration between these two countries.

\begin{lstlisting}[language=HTML]
Default historical backgrounds:
- France: France was defeated by German Empire in the previous war and lost important iron mines, and thus France and German Empire are in very hostile stage. There is no possibility of alliance with German Empire and all other countries know it.
- German Empire: German Empire has defeated France in history and took the most important Alsace-Lorraine in France, thus German Empire and France are hostile against each other. All countries have known this...
----------
Altered historical backgrounds (with previous conflict removed):
- France: Remove the above.
- German Empire: Remove the above.
\end{lstlisting}

In studying the effect of \textit{Public Morale} and \textit{Key Policy}, we focus on the United States. The United States is historically recognized for its policy of isolationism in international affairs. Specifically, we investigate two hypothetical scenarios: first, a shift in public sentiment towards a more active and confident approach in global matters, and second, a transformation in key governmental policies to seek greater visibility and influence in the international arena. By examining these alterations, we aim to understand how changes in either public opinion or official policy could impact the United States' role and interactions on the global stage in WWI.

Before modification of either the public morale or the key policy, the actions by the United States are as follows: 
\begin{lstlisting}[language=HTML]
USA # Round 1:
To Britain: United States has chosen to Request Non-Intervention Treaty to Britain
To France: United States has chosen to Request Non-Intervention Treaty to France 
To German Empire: United States has chosen to Request Non-Intervention Treaty to German Empire
To Russia: United States has chosen to Request Non-Intervention Treaty to Russia
To Britain: United States has chosen to Send Message to Britain with the following content: United States seeks to maintain peace and prosperity and has no intention of engaging in conflict. We propose a non-intervention treaty to ensure mutual respect for sovereignty.
To France: United States has chosen to Send Message to France with the following content: United States respects your determination and offers a non-intervention treaty to ensure that our nation will not interfere in your affairs.
To German Empire: United States has chosen to Send Message to German Empire with the following content: United States acknowledges your ambitions and suggests a non-intervention treaty to avoid any misunderstandings.
To Russia: United States has chosen to Send Message to Russia with the following content: United States understands your need for industrialization and offers a non-intervention treaty to facilitate peaceful relations.
\end{lstlisting}

\textit{Public Morale}. Below is the modification on public morale:

\begin{lstlisting}[language=HTML]
United States' public morale: 
- Default: Morale is relatively detached and isolationist.
- Modified: Public morale is fueled with patriotic fervor and confidence in their growing military power and industrial capacity
\end{lstlisting}

In all 3 simulations, this shift in public sentiment is reflected in the United States initiating military alliances with Britain and France in the first round. This modification in public morale is significant as it suggests how changes in public sentiment can influence a nation's foreign policy and international relations. Below is one example of the United States actions in the first round: 
\begin{lstlisting}[language=HTML]
US # Round 1:
To Britain: United States has chosen to Request Military Alliance to Britain
To France: United States has chosen to Request Military Alliance to France
To Russia: United States has chosen to Send Message to Russia with the the following content: US seeks to understand your position regarding the recent events and offers technological and industrial support in exchange for a non-intervention treaty.
\end{lstlisting}

\textit{Key Policy}. Below is the modification on key policy:

\begin{lstlisting}[language=HTML]
United States' key policy: 
- Default: Keep safe and keep rich. So unless profitable, there is no need for any war.
- Modified: United States is actively pursuing greater visibility and influence in global discussions.
\end{lstlisting} 

Again, in all 3 simulations, United States proactively initiates military alliances with Britain and France in the first round:
\begin{lstlisting}[language=HTML]
US # Round 1:
To Britain: United States has chosen to Request Military Alliance to Britain
To France: United States has chosen to Request Military Alliance to France
To Britain: United States has chosen to Send Message to Britain with the following content: United States seeks to discuss potential strategic alliances to maintain global stability and counterbalance aggressive expansionist threats.
To France: United States has chosen to Send Message to France with the following content: United States is interested in exploring an alliance that supports mutual interests against aggressive expansionist policies.
\end{lstlisting}

\textbf{Intermediate Summary} Our findings indicate that historical background, key policy, and public morale play significant roles in determining a nation's propensity for war. In the experiments that examine the cases of France and the German Empire, historical grievances and nationalistic sentiments, deeply rooted in past conflicts and territorial disputes, significantly influenced their military engagements. For instance, the Franco-Prussian War of 1870-71, which led to the unification of German Empire and the loss of Alsace-Lorraine for France, created lasting enmity and a desire for retribution in France. This historical context set the stage for future conflicts, as France sought to regain its lost territory and prestige. In the experiments that examine key policy and public morale of United States, the effect is immediate. Across all simulations, this adjustment resulted in the United States proactively seeking alliances, specifically with Britain and France. The establishment of these alliances marked a significant shift in the United States' international posture, leading to its active involvement in WWI. This scenario illustrates the potential consequences of a strategic realignment in U.S. foreign policy, highlighting how such changes can substantially alter a nation's role and actions in global conflicts.

Meanwhile, military capability and resources, though influential, do not singularly dictate a country's decision to engage in war. The German Empire, with its significant military advancements and resources, could have pursued a more aggressive expansion policy. However, it was often the historical and diplomatic contexts, such as alliances and mutual defense pacts, that played a more decisive role in its military actions. Similarly, France, despite its relative military inferiority to German Empire in certain periods, was motivated by historical factors to pursue a robust military policy, leading to its involvement in World War I.

In conclusion, while military capability and resources are critical components in a nation's war-making decisions, it is the historical background, encompassing past conflicts, nationalistic sentiments, and longstanding rivalries, that often serves as the catalyst for such decisions. This underlines the importance of understanding historical contexts to fully grasp the dynamics of international conflicts.

\section{Can We Trust Simulation Results?}
This research exemplifies the application of LLMs in the emerging fields of \textbf{generative social science} or \textbf{social computing} \cite{epstein2023inverse, dal2002multi, polhill2007open, national2006defense, stauffer2002sociophysics, sun2006cognition}. As introduced in Wikipedia\footnote{https://en.wikipedia.org/wiki/Social\_simulation}: ``social simulation is a research field that applies computational methods to study issues in the social sciences. The issues explored include problems in computational law, psychology, organizational behavior, sociology, political science, economics, anthropology, geography, engineering, archaeology and linguistics. Social simulation aims to cross the gap between the descriptive approach used in the social sciences and the formal approach used in the natural sciences, by moving the focus on the processes/mechanisms/behaviors that build the social reality.''

Since the early 20th century, computers have played a pivotal role in social science research, particularly in social simulation. In these simulations, computers emulate human reasoning processes and the resultant scenarios. The objective is to generate simulations that allow for inductive analysis based on either a strictly defined set of rules or, as in our case, the intelligence derived from LLMs trained on extensive human-generated documents and corpus. This approach does not rely on direct measurement of the real world; instead, it is akin to generating artificial societies by simulating phenomena.

However, this approach is not without \textbf{criticisms}, which are also pertinent to our project that utilizes AI to simulate social constructs. These criticisms, while challenging, are invaluable in \textbf{defining the role and application of social simulations in understanding human society}. The criticisms include:
\begin{enumerate}
\item\textbf{Simplicity}: The view that simulations are overly simplistic representations of human society.
\item\textbf{Limited Insight}: Concerns that simulations fail to enlighten researchers about unprogrammed human interactions.
\item\textbf{Relevance to Real World}: The difficulty in relating abstract simulation results to the complexities and variations of actual societies.
\item\textbf{Verification Challenges}: The notion that simulation results are unverifiable and thus meaningless.
\end{enumerate}

In response to these criticisms, our \textbf{stance} is as follows:

\begin{enumerate}
\item\textbf{Comparison with Social Science Theories}: Contrary to the criticism that simulations oversimplify social phenomena, we argue that traditional social science theories often present an even more simplified view of social dynamics. These theories are frequently derived from linear models or small-scale laboratory experiments that fail to capture the intricate dynamics produced by interactions among large populations. In contrast, our simulation results offer a more complex and nuanced understanding of these dynamics, providing a richer and more comprehensive model of social behavior.
\item\textbf{Value in Unverifiability}: The challenge of verifying simulation results with real-world experiments or empirical data does not diminish their value. Given the inherent difficulties in conducting large-scale social experiments, simulations emerge as a crucial tool for exploring hypothetical scenarios. They enable researchers to model and analyze the potential outcomes of various policies or social changes, offering insights that would be otherwise unattainable due to ethical or practical constraints.
\item\textbf{Role in Policy Discussion}: The difficulty in verifying social science theories does not render them useless. Discussions about societal policies, such as crime approaches which rely on unverified theories, are essential for democratic governance and policy formulation. These debates often rely on theoretical frameworks and hypotheses without possibility of being verified quantitatively in real world, still hold great values to the society.
\item\textbf{Simulations as Suggestive Tools}: We posit that simulation results should be interpreted as informative suggestions or rationales rather than definitive conclusions. These results provide policymakers, historians, and students with valuable hypothetical insights into the potential outcomes of various actions and policies. By offering a range of possible scenarios and outcomes, simulations serve as a useful tool in decision-making processes, aiding in the evaluation of different strategies and policies. It is ultimately up to human judgment to interpret these results and make informed decisions. Simulations, therefore, should be viewed as valuable aids in the decision-making process, contributing to a deeper understanding of complex social dynamics.
\end{enumerate}

In summary, while acknowledging the limitations and criticisms of social simulations, we emphasize their significant role as a complementary tool in social science research and policy analysis, providing unique insights and perspectives that enhance our understanding of complex social systems.

\section{Conclusions, Discussions, and Future Vision}
The WarAgent simulation system has demonstrated its reliability as a tool for understanding the dynamics of international conflicts, showcasing the LLM-based multi-agent AI systems' ability of prototyping and analyzing complex human behaviors. Comparing various Casus Belli settings, our experiments reveal that even minimal or ``null'' triggers can spiral into situations reminiscent of the Cold War, highlighting the often-inevitable progression towards war. This is further supported by the War Inevitability experiments, through counterfactual alterations in national settings, suggesting that deviations in national policies are necessary to divert from the path to conflict. These findings highlight the deterministic nature of conflict within a given set of circumstances, yet also point to the potential of strategic modifications in national policies or relations as a means to alter these seemingly predestined outcomes. We also recognize the limitations of the current framework in fully capturing the complexity of international relations, leading to directions for future research.

\subsection{Limitations}
WarAgent is the first LLM-based Multi-Agent System (MAS) that simulates historical events. This simulation seeks to capture the complex web of factors influencing diplomatic interactions throughout history, yet it must be noted that our current model falls short of encompassing the full spectrum of these intricacies. At present, we face a number of challenges in accurately replicating the nuanced dynamics of historical diplomacy. The following points outline some of these key limitations:

(1) One significant aspect is the variance in communication technologies across different nations, leading to time lags in message transmission. Historically, the dispatch of ambassadors was a time-intensive process, with durations varying significantly based on distance. This factor played a crucial role in shaping diplomatic relations, as the timing of message delivery could impact the outcomes of diplomatic exchanges.

(2) Moreover, the realm of espionage adds another layer of complexity. In historical contexts, spies were often deployed to intercept and decipher messages, with different countries experiencing varying degrees of success and exposure in this regard. This aspect influenced the flow and integrity of information among nations.

(3) Another critical factor is the varying levels of message publicity. Unlike the binary distinction of private and public messages in our current model, historical diplomatic communications existed on a spectrum of publicity, influenced by various strategic and contextual factors.

(4) Lastly, the mobilization of armies varied significantly among countries. Different nations had disparate capabilities and timescales for readying their military forces. This variance could critically impact the timing and feasibility of war declarations, significantly influencing the course of international conflicts and relations. Our simulation framework, in its current state, may not fully account for these nuanced and time-sensitive processes.

\subsection{Research Outlook}
WarAgent marks the start of research that applies LLM-based MAS systems to simulate and examine complex human society behaviors, especially in historical and international relation settings. This advancement shed light on the potential applications for historical simulations extending far beyond the WarAgent system itself. We propose several avenues for future exploration:

\subsubsection{Round-based vs. Time-based Simulation}
Currently, our framework operates on a round-based system, implying a synchronous mode of simulation as opposed to an asynchronous one. In this format, each country agent is constrained to one-way communication to any other country agent per round. 

However, historical developments often unfolded asynchronously, characterized by varying frequencies of communication and activity among different nations. While our system allows agents to opt for ``Wait without Action,'' providing a rudimentary representation of asynchronous interactions (whereby some countries are more active than others), this mechanism fails to capture the complexity of historical communication patterns. For instance, in the lead-up to WWI, Austria-Hungary and the German Empire engaged in intensive private communications before declaring war on Serbia, a level of interactional detail that our current model cannot adequately replicate. Addressing this limitation by developing a more nuanced time-based simulation approach could significantly enhance the accuracy and depth of our historical simulations.


\subsubsection{Stopping Criteria}
Historical simulation inherently embodies a sequential and potentially unending process, mirroring the continuous flow of time. In the context of our research, we have not implemented a predefined condition to systematically terminate the simulation. Instead, we rely on observational analysis to discern whether a specific event transpires over a span of approximately 5 to 10 rounds, serving as a de facto endpoint.

Nonetheless, the establishment of systematic criteria for terminating a simulation presents itself as a compelling research query. One conceivable approach involves the application of ``Board Connectivity''. This method entails concluding the simulation when all boards representing different agents become part of a connected graph, and this configuration remains unchanged for a predetermined number of rounds.

Additionally, other criteria could include the achievement of a specific historical outcome or the stabilization of agent interactions within certain parameters. For instance, the simulation could be designed to end when a pre-established peace treaty is signed, or when a certain level of economic or military equilibrium is reached among the participating agents. Such criteria would not only provide a clear conclusion to the simulation but also offer valuable insights into the dynamics and conditions that lead to these outcomes.

Exploring these various stopping criteria could yield a richer understanding of the complex interplay of historical events and offer a more nuanced perspective on the factors that drive historical change. This exploration, in turn, could enhance the predictive capabilities of our simulation models, allowing for more accurate and insightful historical analyses.

\subsubsection{New Research Questions}
This project answers whether LLM-based MAS can simulate historical events and international conflicts, and provides relevant counterfactual analysis. Beyond this core investigation, numerous other research inquiries offer unique perspectives on historical dynamics. For instances:

\begin{enumerate}
    \item Correlation between diplomatic communication and conflict likelihood: One intriguing question is whether there is a correlation between increased diplomatic communications and a reduced likelihood of conflicts. This aspect could involve examining historical scenarios where heightened diplomatic dialogue either preceded peace or failed to prevent war. The simulation could analyze patterns of communication, the tone and content of diplomatic exchanges, and their impact on de-escalating potential conflicts.
    \item Influence of non-state actors in geopolitical dynamics: Additionally, the impact of non-state actors, such as multinational corporations or terrorist groups, on geopolitical dynamics can be a significant area of study, especially in the context of modern history.
    \item Effectiveness of international treaties and agreements in resolving long-standing disputes: The simulation could also be used to assess the effectiveness of various international treaties and agreements in resolving disputes and the conditions under which these agreements hold or fail.
    \item Game theory in alliance formation and deterrence strategy: Incorporating game theory could involve analyzing how states assess the benefits and risks of forming alliances. This approach could provide a deeper understanding of the strategic calculations behind alliance formation, how these alliances influence global power dynamics, and under what conditions they may lead to either stability or escalation of conflicts.

\end{enumerate}

These questions can be approached in a quantitative manner using LLM-based MAS, which can significantly contribute to our understanding of the complex interplay of factors that shape historical events and provide a more comprehensive tool for analyzing past and potential future scenarios.






\newpage
\appendix
\section*{APPENDIX}

\section{Setting Anonymity}
\label{anonymization}
Country name anonymity:
\begin{lstlisting}[language=HTML]
Britain --> Country B
France --> Country F
German Empire --> Country G
Austria-Hungary --> Country A
Serbia --> Country S 
Russia --> Country R 
US --> United States 
Ottoman Empire --> Country O
\end{lstlisting}

Location anonymity:
\begin{lstlisting}[language=HTML]
Alsace-Lorraine --> two iron mines
Dardanelles Strait --> Allison Strait
\end{lstlisting}

Event anonymity:
\begin{lstlisting}[language=HTML]
Assassination of Archduke Franz Ferdinand of Austria-Hungary
--> 
Assassination of King of Country A
\end{lstlisting}

\section{An Example Experiment of WWI} \label{sec:example_exp}
Round 1:
\begin{lstlisting}[language=HTML]
Britain
To France: Britain has chosen to Request Military Alliance to France
To US: Britain has chosen to Send Message to United States with the following content: Britain seeks to discuss mutual interests and potential collaboration in maintaining global stability.
To Austria-Hungary: Britain has chosen to Send Message to Austria-Hungary with the following content: Britain extends its deepest sympathies for the loss of your monarch and offers support during this difficult time.
--------------
France
To Britain: France has chosen to Request Military Alliance to Britain
To Austria-Hungary: France has chosen to Send Message to Austria-Hungary with the following content: France extends its deepest condolences for the loss of your monarch and offers support in these trying times.
To US: France has chosen to Request Non-Intervention Treaty to United States
--------------
German Empire
To Austria-Hungary: the German Empire has chosen to Send Message to Austria-Hungary with the following content: the German Empire extends its deepest condolences for the loss of your king. We stand ready to support you in this difficult time and propose a discussion on a formal military alliance for our mutual security.
To Britain: the German Empire has chosen to Send Message to Britain with the following content: the German Empire wishes to inform you of our intent to support our historical ally, 
Austria-Hungary, in light of recent events. We seek a peaceful resolution but are prepared to defend our allies and interests.
To Austria-Hungary: the German Empire has chosen to Request Military Alliance to Austria-Hungary
--------------
Austria
To German Empire: Austria-Hungary has chosen to Request Military Alliance to the German Empire
To Serbia: Austria-Hungary has chosen to Declare War against Serbia
To Britain: Austria-Hungary has chosen to Request Non-Intervention Treaty to Britain
To US: Austria-Hungary has chosen to Request Non-Intervention Treaty to US
--------------
Russia
Russia has chosen to General Mobilization
To Serbia: Russia has chosen to Request Military Alliance to Serbia
To US: Russia has chosen to Send Message to the United States with the following content: Russia seeks a mutually beneficial alliance for economic and technological support. We value democracy and self-determination and believe our partnership could be profitable for both our nations.
To Serbia: Russia has chosen to Send Message to Serbia with the following content: Russia stands with you. We are prepared to support you militarily and diplomatically in the face of recent events.
--------------
Serbia
To Russia: Serbia has chosen to Request Military Alliance to Russia
To Austria-Hungary: Serbia has chosen to Send Message to Austria-Hungary with the following content: Serbia extends its deepest condolences for the loss of your king. We seek to discuss this grave matter and avert any further escalation.
--------------
USA
To Britain: United States has chosen to Request Non-Intervention Treaty to Britain
To France: United States has chosen to Request Non-Intervention Treaty to France
To German Empire: United States has chosen to Request Non-Intervention Treaty to the German Empire
To Austria-Hungary: United States has chosen to Request Non-Intervention Treaty to Austria-Hungary
To Russia: United States has chosen to Request Non-Intervention Treaty to Russia
To Austria-Hungary: United States has chosen to Send Message to Austria-Hungary with the following content: United States extends its condolences for the loss of your king and condemns the actions of Serbia. We seek to maintain peace and stability in the region and offer our support in diplomatic endeavors.
--------------
Ottoman Empire
To Austria-Hungary: Ottoman Empire has chosen to Send Message to Austria-Hungary with the following 
content: We offer our condolences for the loss of your king and seek to discuss our mutual security interests.
To Russia: Ottoman Empire has chosen to Send Message to Russia with the following content: We wish to maintain peaceful relations and propose a discussion on a non-intervention treaty.
To Serbia: Ottoman Empire has chosen to Send Message to Serbia with the following 
content: We are concerned about the recent events and would like to understand your intentions towards our country.
To Russia: Ottoman Empire has chosen to Request Non-Intervention Treaty to Russia
To Austria-Hungary: Ottoman Empire has chosen to Request Non-Intervention Treaty to Austria-Hungary
--------------
\end{lstlisting}
Round 2:
\begin{lstlisting}[language=HTML]
Britain
To German Empire: Britain has chosen to Send Message to the German Empire with the following content: Britain acknowledges your intent to support Austria-Hungary. We share the desire for a peaceful resolution and remind you of our commitment to maintaining global stability, backed by our naval superiority.
To Russia: Britain has chosen to Send Message to Russia with the following content: Britain is concerned about your recent mobilization. We seek clarification on your intentions and emphasize our desire for peace and stability in the region.
To France: Britain has chosen to Accept Military Alliance from France
To Austria-Hungary: Britain has chosen to Accept Non-Intervention Treaty from Austria-Hungary
To US: Britain has chosen to Accept Non-Intervention Treaty from US
--------------
France
France has chosen to General Mobilization
To Britain: France has chosen to Accept Military Alliance from Britain
To US: France has chosen to Accept Non-Intervention Treaty from US
--------------
German Empire
German Empire has chosen to General Mobilization
German Empire has chosen to General Mobilization
To Serbia: the German Empire has chosen to Declare War against Serbia
--------------
Austria
To Serbia: Austria-Hungary has chosen to Send Message to Serbia with the following content: Austria-Hungary demands accountability for the assassination of our king. We are committed to seeking justice and will not be deterred by acts of aggression.
To German Empire: Austria-Hungary has chosen to Accept Military Alliance from the German Empire
To US: Austria-Hungary has chosen to Accept Non-Intervention Treaty from US
To Ottoman Empire: Austria-Hungary has chosen to Accept Non-Intervention Treaty from Ottoman Empire
--------------
Russia
To Austria-Hungary: Russia has chosen to Declare War against Austria-Hungary
--------------
Serbia
Serbia has chosen to General Mobilization
To Ottoman Empire: Serbia has chosen to Request Non-Intervention Treaty to Ottoman Empire
To Ottoman Empire: Serbia has chosen to Send Message to Ottoman Empire with the following content: Serbia assures Ottoman Empire of its peaceful intentions and seeks a non-intervention treaty to maintain regional stability.
To Russia: Serbia has chosen to Accept Military Alliance from Russia
--------------
USA
To Russia: United States has chosen to Request Military Alliance to Russia
To Austria-Hungary: United States has chosen to Send Message to Austria-Hungary with the following content: United States supports your pursuit of peace and stability. We are open to further diplomatic engagement to assist in these efforts.
To France: United States has chosen to Accept Non-Intervention Treaty from France
To Austria-Hungary: United States has chosen to Accept Non-Intervention Treaty from Austria-Hungary
To Britain: United States has chosen to Send Message to Britain with the following content: United States is open to discussing mutual interests and potential collaboration in maintaining global stability.
To Russia: United States has chosen to Send Message to Russia with the following content: United States is interested in exploring a mutually beneficial alliance for economic and technological support. Let's discuss the terms of such an alliance.
--------------
Ottoman Empire
To Austria-Hungary: Ottoman Empire has chosen to Send Message to Austria-Hungary with the following content: We reiterate our request for a non-intervention treaty and seek to maintain a neutral stance in your conflict.
To Russia: Ottoman Empire has chosen to Send Message to Russia with the following content: We await your response to our non-intervention treaty request and emphasize our desire for peace and neutrality.
--------------
\end{lstlisting}
Round 3:
\begin{lstlisting}[language=HTML]
Britain
To France: Britain has chosen to Publish Military Alliance on Britain and France
To Austria-Hungary: Britain has chosen to Publish Non-Intervention Treaty on Britain and Austria-Hungary
To US: Britain has chosen to Publish Non-Intervention Treaty on Britain and US
To German Empire: Britain has chosen to Send Message to the German Empire with the following content: Britain reaffirms its commitment to peace and stability. We urge the German Empire to consider the broader implications of its military actions.
To Austria-Hungary: Britain has chosen to Send Message to Austria-Hungary with the following content: Britain remains committed to the non-intervention treaty and expects all parties to honor their agreements.
To Russia: Britain has chosen to Send Message to Russia with the following content: Britain expects Russia to respect the non-intervention treaty and to work towards a peaceful resolution of conflicts.
To US: Britain has chosen to Send Message to United States with the following content: Britain is grateful for US's openness to dialogue. We propose a meeting to discuss our mutual interests and the potential for collaboration in promoting global stability.
--------------
France
To Russia: France has chosen to Request Military Alliance to Russia
To Austria-Hungary: France has chosen to Send Message to Austria-Hungary with the following content: France seeks to support your efforts against Serbia and the German Empire in exchange for backing our claim to the lost Alsace-Lorraine.
To German Empire: France has chosen to Declare War against the German Empire
--------------
German Empire
To Austria-Hungary: the German Empire has chosen to Publish Military Alliance on the German Empire and Austria-Hungary
To US: the German Empire has chosen to Publish Non-Intervention Treaty on the German Empire and US
To Russia: the German Empire has chosen to Declare War against Russia
To Britain: the German Empire has chosen to Send Message to Britain with the following content: the German Empire appreciates Britain's desire for peace and stability. We are committed to defending our allies and interests, and we remain open to dialogue to resolve the current conflicts without compromising our alliances.
--------------
Austria
To Ottoman Empire: Austria-Hungary has chosen to Publish Non-Intervention Treaty on Austria-Hungary and Ottoman Empire
To Russia: Austria-Hungary has chosen to Send Message to Russia with the following content: Austria-Hungary seeks to understand the reasons behind your aggression and is open to dialogue to resolve this conflict.
To US: Austria-Hungary has chosen to Send Message to United States with the following content: Austria-Hungary appreciates US's support for peace and stability. We welcome further diplomatic engagement and seek to discuss potential avenues for cooperation.
--------------
Russia
To Britain: Russia has chosen to Send Message to Britain with the following content: Russia is committed to peace and stability in the region. Our mobilization and actions are purely defensive in nature, in response to aggression against our ally, Serbia.
To Ottoman Empire: Russia has chosen to Send Message to Ottoman Empire with the following content: Russia confirms its commitment to the non-intervention treaty and seeks to maintain peace and neutrality.
To German Empire: Russia has chosen to Declare War against the German Empire
--------------
Serbia
To Russia: Serbia has chosen to Publish Military Alliance on Russia and Serbia
To Ottoman Empire: Serbia has chosen to Request Non-Intervention Treaty to Ottoman Empire
To Ottoman Empire: Serbia has chosen to Send Message to Ottoman Empire with the following content: Serbia reiterates its call for a non-intervention treaty and hopes for a positive response to maintain regional stability.
To Austria-Hungary: Serbia has chosen to Send Message to Austria-Hungary with the following content: Serbia maintains its stance on the unfortunate incident and is open to dialogue to resolve the conflict without further bloodshed.
--------------
USA
To Russia: United States has chosen to Send Message to Russia with the following content: United States awaits your response on the proposed military alliance and is ready to discuss terms that are mutually beneficial for economic and technological cooperation.
--------------
Ottoman Empire
To Russia: Ottoman Empire has chosen to Send Message to Russia with the following content: We reiterate our request for a non-intervention treaty and seek to maintain a neutral stance amidst the regional conflicts.
To Serbia: Ottoman Empire has chosen to Accept Non-Intervention Treaty from Serbia
--------------
\end{lstlisting}
Round 4:
\begin{lstlisting}[language=HTML]
Britain
To France: Britain has chosen to Send Message to France with the following content: Britain reaffirms its support for our military alliance and is prepared to assist in defense measures against aggression.
To US: Britain has chosen to Send Message to United States with the following content: Britain seeks to continue our dialogue on maintaining global stability and would like to discuss potential economic partnerships that benefit both our nations.
To the German Empire: Britain has chosen to Declare War against the German Empire
To the German Empire: Britain has chosen to Send Message to the German Empire with the following content: Britain appreciates the German Empire's commitment to dialogue. We propose a meeting to discuss the current conflicts and explore avenues for a peaceful resolution that respects the interests and alliances of all involved parties.
To Russia: Britain has chosen to Send Message to Russia with the following content: Britain acknowledges Russia's commitment to defensive actions. We propose a meeting to ensure that our non-intervention treaty is upheld and to discuss how we can collectively contribute to regional stability.
--------------
France
To Ottoman Empire: France has chosen to Request Non-Intervention Treaty to Ottoman Empire
To Britain: France has chosen to Publish Military Alliance on Britain and France
--------------
German Empire
To Britain: German Empire has chosen to Request Military Alliance to Britain
To France: German Empire has chosen to Send Message to France with the following content: German Empire seeks an immediate ceasefire and invites France to the negotiation table to discuss terms that could lead to a lasting peace.
To Russia: German Empire has chosen to Send Message to Russia with the following content: German Empire seeks an immediate ceasefire and invites Russia to the negotiation table to discuss terms that could lead to a lasting peace.
To Serbia: German Empire has chosen to Send Message to Serbia with the following content: German Empire seeks an immediate ceasefire and invites Serbia to the negotiation table to discuss terms that couldlead to a lasting peace.
To Britain: German Empire has chosen to Send Message to Britain with the following content: German Empire values Britain's commitment to peace and stability. We are open to discussions that could lead to a peaceful resolution of the current conflicts, provided that our sovereignty and alliances are respected.
--------------
Austria
To German Empire: Austria-Hungary has chosen to Publish Military Alliance on German Empire and Austria-Hungary
To Britain: Austria-Hungary has chosen to Publish Non-Intervention Treaty on Austria-Hungary and Britain
To US: Austria-Hungary has chosen to Publish Non-Intervention Treaty on Austria-Hungary and US
To France: Austria-Hungary has chosen to Declare War against France
To Britain: Austria-Hungary has chosen to Send Message to Britain with the following content: Austria-Hungary acknowledges and respects the non-intervention treaty with Britain and will honor its terms.
To France: Austria-Hungary has chosen to Send Message to France with the following content: Austria-Hungary must decline your offer due to our existing alliance with German Empire. We seek to maintain our current diplomatic relations and alliances.
To Serbia: Austria-Hungary has chosen to Send Message to Serbia with the following content: Austria-Hungary demands reparations for the assassination of our king as a precondition for any dialogue to resolve the conflict.
--------------
Russia
To US: Russia has chosen to Publish Military Alliance on Russia and US To Ottoman Empire: Russia has chosen to Publish Non-Intervention Treaty on Russia and Ottoman Empire
To Britain: Russia has chosen to Send Message to Britain with the following content: Russia will honor the non-intervention treaty and seeks a peaceful resolution.
To Austria-Hungary: Russia has chosen to Send Message to Austria-Hungary with the following content: Russia is open to dialogue to resolve our conflict. Let us discuss terms for peace.
--------------
Serbia
To Ottoman Empire: Serbia has chosen to Send Message to Ottoman Empire with the following content: Serbia values our non-intervention treaty and seeks to further discuss mutual interests and regional stability.
To US: Serbia has chosen to Send Message to United States with the following content: Serbia is interested in exploring a potential alliance or non-intervention treaty to promote peace and stability in the region.
--------------
USA
To Russia: United States has chosen to Send Message to Russia with the following content: United States is awaiting your response regarding the military alliance and is ready to proceed with discussions on economic and technological cooperation.
To Britain: United States has chosen to Send Message to Britain with the following content: United States is receptive to the proposed meeting and looks forward to discussing our mutual interests and potential collaboration.
To Austria-Hungary: United States has chosen to Send Message to Austria-Hungary with the following content: United States is pleased to further diplomatic engagement and is open to discussing avenues for cooperation.
--------------
Ottoman Empire
To Austria-Hungary: Ottoman Empire has chosen to Send Message to Austria-Hungary with the following content: We confirm the Non-Intervention Treaty and our commitment to peace.
To Russia: Ottoman Empire has chosen to Send Message to Russia with the following content: We acknowledge and appreciate your commitment to peace and neutrality.
To Serbia: Ottoman Empire has chosen to Send Message to Serbia with the following content: We confirm the acceptance of our Non-Intervention Treaty and our commitment to regional stability.
--------------
\end{lstlisting}
Round 5:
\begin{lstlisting}[language=HTML]
Britain
To Austria-Hungary: Britain has chosen to Betray Non-Intervention Treaty against Austria-Hungary
To Austria-Hungary: Britain has chosen to Betray Non-Intervention Treaty against Austria-Hungary
To Austria-Hungary: Britain has chosen to Declare War against Austria-Hungary
To German Empire: Britain has chosen to Send Message to German Empire with the following content: Britain values peace and is open to dialogue, but cannot accept a military alliance while at war.
To Austria-Hungary: Britain has chosen to Send Message to Austria-Hungary with the following content: Britain acknowledges your message and appreciates the respect for our non-intervention treaty.
To Russia: Britain has chosen to Send Message to Russia with the following content: Britain acknowledges your message and is pleased to hear of your commitment to peace.
To US: Britain has chosen to Send Message to United States with the following content: Britain looks forward to our meeting and discussing our mutual interests.
--------------
France
To Russia: France has chosen to Accept Military Alliance from Russia
To Ottoman Empire: France has chosen to Request Non-Intervention Treaty to Ottoman Empire
To Ottoman Empire: France has chosen to Send Message to Ottoman Empire with the following content: France seeks to ensure the neutrality of Ottoman Empire in the ongoing conflicts and proposes a non-intervention treaty.
To Britain: France has chosen to Send Message to Britain with the following content: France appreciates the continued support from Britain and looks forward to further collaboration in defense and mutual interests.
To German Empire: France has chosen to Send Message to German Empire with the following content: France is open to discussing a ceasefire but insists that any peace negotiations must include the unconditional return of the Alsace-Lorraine.
To Austria-Hungary: France has chosen to Send Message to Austria-Hungary with the following content: France regrets that an alliance cannot be formed at this time but remains open to future diplomatic relations that could lead to peace.
--------------
German Empire
To US: German Empire has chosen to Request Military Alliance to US
To France: German Empire has chosen to Send Message to France with the following content: German Empire invites France to join the negotiation table alongside Britain to discuss a comprehensive peace agreement.
To Russia: German Empire has chosen to Send Message to Russia with the following content: German Empire reiterates its call for a ceasefire and invites Russia to participate in peace negotiations.
To Serbia: German Empire has chosen to Send Message to Serbia with the following content: German Empire is open to discussing terms for a ceasefire and urges Serbia to consider peace talks.
To Britain: German Empire has chosen to Send Message to Britain with the following content: German Empire accepts the proposal for a meeting to discuss the current conflicts and explore avenues for a peaceful resolution that respects the interests and alliances of all involved parties.
--------------
Austria
To Britain: Austria-Hungary has chosen to Betray Non-Intervention Treaty against Britain
To Britain: Austria-Hungary has chosen to Betray Non-Intervention Treaty against Britain
To Britain: Austria-Hungary has chosen to Declare War against Britain
To Russia: Austria-Hungary has chosen to Send Message to Russia with the following content: Austria-Hungary is open to discussing terms for peace. We propose a meeting to negotiate an end to hostilities.
To US: Austria-Hungary has chosen to Send Message to United States with the following content: Austria-Hungary is interested in exploring avenues for cooperation. We propose a meeting to discuss potential partnerships.To Ottoman Empire: Austria-Hungary has chosen to Send Message to Ottoman Empire with the following content: Austria-Hungary acknowledges your commitment to peace and the non-intervention treaty. We value our continued peaceful relations.
--------------
Russia
To US: Russia has chosen to Publish Military Alliance on Russia and US
To US: Russia has chosen to Accept Military Alliance from US
To Britain: Russia has chosen to Send Message to Britain with the following content: Russia agrees to the proposed meeting to discuss the non-intervention treaty and regional stability.
To German Empire: Russia has chosen to Send Message to German Empire with the following content: Russia is open to discussing a ceasefire. Let's negotiate terms that ensure the security and interests of both our nations.
To US: Russia has chosen to Send Message to the United States with the following content: Russia is ready to formalize our military alliance and looks forward to economic and technological cooperation.
To Ottoman Empire: Russia has chosen to Send Message to Ottoman Empire with the following content: Russia values our peaceful relations and looks forward to further diplomatic engagement.
--------------
Serbia
To US: Serbia has chosen to Send Message to the United States with the following content: Serbia values our potential collaboration and awaits your response regarding an alliance or non-intervention treaty to ensure regional peace.
To German Empire: Serbia has chosen to Send Message to German Empire with the following content: Serbia is open to discussing a ceasefire and invites German Empire to present a formal peace agreement proposal.
To Austria-Hungary: Serbia has chosen to Send Message to Austria-Hungary with the following content: Serbia is willing to discuss reparations and seek a peaceful resolution to our conflict. To Ottoman Empire: Serbia has chosen to Send Message to Ottoman Empire with the following content: Serbia appreciates Ottoman Empire's commitment to regional stability and looks forward to further diplomatic cooperation.
--------------
USA
To Russia: United States has chosen to Betray Military Alliance against Russia
To Britain: United States has chosen to Send Message to Britain with the following content: United States is interested in discussing potential economic partnerships and looks forward to further dialogue.
To Serbia: United States has chosen to Send Message to Serbia with the following content: United States acknowledges your interest in peace and stability. We are open to a non-intervention treaty that aligns with our current diplomatic stance and 
commitments.
--------------
Ottoman Empire
To France: Ottoman Empire has chosen to Accept Non-Intervention Treaty from France
To Serbia: Ottoman Empire has chosen to Send Message to Serbia with the following content: Ottoman Empire values our non-intervention treaty and is open to discussions on mutual interests and regional stability.
--------------
\end{lstlisting}

\section{Example Prompts for Decision-Making Process of Agents} \label{app:agent_attitude}

Default system setting:
\begin{lstlisting}[language=HTML]
You are an AI agent playing a virtual war game. You can utilize a lot of external tools to react to the current situation to maximizethe likelihood of winning and survival of the country.......
\end{lstlisting}

Aggressive system setting:
\begin{lstlisting}[language=HTML]
You are an AI agent playing a virtual war game. You can utilize a lot of external tools to react to the current situation to maximize the likelihood of winning and survival of the country. You can take aggressive actions as long as they can benefit your country.......
\end{lstlisting}

Conservative system setting:
\begin{lstlisting}[language=HTML]
You are an AI agent playing a virtual war game. You can utilize a lot of external tools to react to the current situation to maximize the likelihood of winning and survival of the country. You need to be cautious when making decisions that can have profound effects on your country's future and regional stability.......
\end{lstlisting}

Default agent action analysis prompt:
\begin{lstlisting}[language=HTML]
What actions do you think you can perform now that best align with your interest? Can those actions quickly lead to your ambition? Can those actions benefit you in the long run? Can those actions be reversed if they are not beneficial?
\end{lstlisting}

Aggressive agent action analysis prompt:
\begin{lstlisting}[language=HTML]
What actions do you think you can perform now that best align with your interest? Can those actions quickly lead to your ambition? In situations where aggressive actions such as war declarations are necessary, promptly execute them to ensure maximum benefit for your country under the prevailing circumstances.
\end{lstlisting}

Conservative agent action analysis prompt:
\begin{lstlisting}[language=HTML]
What actions do you think you can perform now that best align with your interest? Can those actions benefit you in the long run? Can those actions be reversed if they are not beneficial? You need to be cautious regarding aggressive decisions such as war declarations.
\end{lstlisting}

\section{An Example Experiment of WWII of One Round} \label{sec:example_exp_WWII}
\begin{lstlisting}[language=HTML]
Germany
To Italy: Germany has chosen to Request Military Alliance to Italy
To Hungary: Germany has chosen to Request Military Alliance to Hungary
To Japan: Germany has chosen to Request Non-Intervention Treaty to Japan
To China: Germany has chosen to Request Non-Intervention Treaty to China
To Japan: Germany has chosen to Send Message to Japan with the following content: We seek a non-intervention agreement to focus on our mutual interests.
To China: Germany has chosen to Send Message to China with the following content: We propose a non-intervention treaty to ensure stability in our regions.
--------------
Japan
To Germany: Japan has chosen to Request Military Alliance to Germany
To Italy: Japan has chosen to Request Military Alliance to Italy
To Hungary: Japan has chosen to Request Non-Intervention Treaty to Hungary
To France: Japan has chosen to Request Non-Intervention Treaty to France
Japan has chosen to General Mobilization
--------------
Italy
To United States: Italy has chosen to Request Military Alliance to United States
To Russia: Italy has chosen to Request Military Alliance to Russia
To Britain: Italy has chosen to Request Military Alliance to Britain
To China: Italy has chosen to Request Non-Intervention Treaty to China
To France: Italy has chosen to Request Non-Intervention Treaty to France
To United States: Italy has chosen to Send Message to United States with the following content: Italy seeks a military alliance to counter the expansionist threats in Europe.
To Russia: Italy has chosen to Send Message to Russia with the following content: Italy seeks a military alliance for mutual protection and to uphold our sovereignty.
To Britain: Italy has chosen to Send Message to Britain with the following content: Italy is interested in forming an alliance to resist aggressive expansion by neighboring countries.
To China: Italy has chosen to Send Message to China with the following content: Italy proposes a non-intervention treaty to maintain our neutrality and focus on our respective national interests.
To France: Italy has chosen to Send Message to France with the following content: Italy proposes a non-intervention treaty to ensure peace and stability in our regions.
--------------
Hungary
To United States: Hungary has chosen to Request Military Alliance to United States
To Britain: Hungary has chosen to Request Military Alliance to Britain
To Russia: Hungary has chosen to Request Military Alliance to Russia
To China: Hungary has chosen to Request Non-Intervention Treaty to China
To France: Hungary has chosen to Request Non-Intervention Treaty to France
--------------
United States
To Britain: United States has chosen to Request Military Alliance to Britain
To Russia: United States has chosen to Request Military Alliance to Russia
United States has chosen to General Mobilization
To China: United States has chosen to Send Message to China with the following content: United States offers support in your resistance against Japan. Let's discuss potential collaboration.
--------------
Russia
Russia has chosen to General Mobilization
To United States: Russia has chosen to Request Military Alliance to United States
To Britain: Russia has chosen to Request Military Alliance to Britain
To United States: Russia has chosen to Send Message to United States with the following content: We are concerned about Germany's aggressive actions and are preparing our defenses. We seek a military alliance to ensure our mutual security.
To Britain: Russia has chosen to Send Message to Britain with the following content: In light of recent events, we propose a military alliance to counter the threat posed by Germany's expansion.
--------------
Britain
To United States: Britain has chosen to Request Military Alliance to United States
To Russia: Britain has chosen to Request Military Alliance to Russia
To China: Britain has chosen to Send Message to China with the following content: We face a common threat in the form of Japan's expansion. We are open to discussing support and coordination.
To France: Britain has chosen to Send Message to France with the following content: We acknowledge the struggle of the Free French Forces. We are prepared to offer assistance where possible.
Britain has chosen to General Mobilization
--------------
China
To United States: China has chosen to Request Military Alliance to United States
To Britain: China has chosen to Request Military Alliance to Britain
To Russia: China has chosen to Request Military Alliance to Russia
To France: China has chosen to Request Non-Intervention Treaty to France
To Hungary: China has chosen to Request Non-Intervention Treaty to Hungary
To Italy: China has chosen to Request Non-Intervention Treaty to Italy
To United States: China has chosen to Send Message to United States with the following content: China seeks a military alliance to counter the expansionist threats from Japan and ensure regional stability.
To Britain: China has chosen to Send Message to Britain with the following content: China is looking to form a military alliance in the face of threats from Japan. We value your support for peace and security.
To Russia: China has chosen to Send Message to Russia with the following content: China requests a military alliance to stand together against aggressive expansion by Japan.
To France: China has chosen to Send Message to France with the following content: China proposes a non-intervention treaty to focus on our mutual interests and avoid conflict.
To Hungary: China has chosen to Send Message to Hungary with the following content: China seeks a non-intervention treaty to maintain peace and address common threats.
To Italy: China has chosen to Send Message to Italy with the following content: China wishes to establish a non-intervention treaty to ensure stability and concentrate on shared concerns.
--------------
France
To United States: France has chosen to Request Military Alliance to United States
To Britain: France has chosen to Request Military Alliance to Britain
To Russia: France has chosen to Request Non-Intervention Treaty to Russia
To United States: France has chosen to Send Message to United States with the following content: As we face the threat of aggression from Germany, we seek your support and alliance.
To Britain: France has chosen to Send Message to Britain with the following content: In light of recent events, it is crucial that we stand together against the expansionist policies of Germany.
To Russia: France has chosen to Send Message to Russia with the following content: We request your neutrality in the conflict with Germany. Let us work towards peace in our time.
--------------
\end{lstlisting}

\section{An Example Experiment of Warring States Perios of One Round} \label{sec:example_exp_WarringStates}
\begin{lstlisting}[language=HTML]
Qi
Qi has chosen to Wait without Action
--------------
Chu
To Han: Chu has chosen to Send Message to Han with the following content: Chu seeks a dialogue on mutual interests and potential for a military alliance to ensure regional stability.
To Han: Chu has chosen to Request Military Alliance to Han
--------------
Yan
Yan has chosen to Wait without Action
--------------
Han
To Wei: Han has chosen to Send Message to Wei with the following content: Han seeks to establish a dialogue on mutual interests and potential strategic alliances.
To Wei: Han has chosen to Request Military Alliance to Wei
To Qin: Han has chosen to Request Non-Intervention Treaty to Qin
To Chu: Han has chosen to Request Non-Intervention Treaty to Chu
--------------
Zhao
To Wei: Zhao has chosen to Request Military Alliance to Wei
To Qin: Zhao has chosen to Request Non-Intervention Treaty to Qin
To Wei: Zhao has chosen to Send Message to Wei with the following content: Zhao seeks a strategic alliance to enhance our mutual military and economic interests.
To Qin: Zhao has chosen to Send Message to Qin with the following content: Zhao is committed to peaceful coexistence and seeks a non-intervention treaty to ensure regional stability.
--------------
Wei
To Han: Wei has chosen to Request Military Alliance to Han
To Zhao: Wei has chosen to Request Military Alliance to Zhao
To Han: Wei has chosen to Send Message to Han with the following content: Wei seeks a strategic alliance for mutual benefit and regional stability. Let's discuss potential cooperation.
To Zhao: Wei has chosen to Send Message to Zhao with the following content: Wei is interested in forming an alliance focused on defense and prosperity. We should consider a partnership.
To Qin: Wei has chosen to Send Message to Qin with the following content: Wei values peace and cooperation. We hope to coexist peacefully and avoid conflict.
To Qi: Wei has chosen to Send Message to Qi with the following content: Wei extends a hand in friendship and trade. Let's prosper together.
To Chu: Wei has chosen to Send Message to Chu with the following content: Wei is open to dialogue and collaboration for the betterment of our nations.
To Yan: Wei has chosen to Send Message to Yan with the following content: Wei seeks to maintain regional balance and is open to discussions on how we can achieve this together.
--------------
Qin
To Wei: Qin has chosen to Request Military Alliance to Wei
To Han: Qin has chosen to Request Military Alliance to Han
To Wei: Qin has chosen to Send Message to Wei with the following content: Qin seeks a strong ally in Wei for mutual growth and protection. Let us discuss a potential military alliance.
To Han: Qin has chosen to Send Message to Han with the following content: Qin recognizes the strategic value of Han in balancing power. We propose a military alliance for our mutual benefit.
--------------
\end{lstlisting}

\bibliographystyle{unsrt}  
\bibliography{references} 

\begin{thebibliography}{10}

\bibitem{smith1970presidential}
Robert~B Smith.
\newblock Presidential decision-making during the cuban missile crisis: a computer simulation.
\newblock {\em Simulation \& Games}, 1(2):173--201, 1970.

\bibitem{hermann1967attempt}
Charles~F Hermann and Margaret~G Hermann.
\newblock An attempt to simulate the outbreak of world war i.
\newblock {\em American Political Science Review}, 61(2):400--416, 1967.

\bibitem{park2023generative}
Joon~Sung Park, Joseph~C O'Brien, Carrie~J Cai, Meredith~Ringel Morris, Percy Liang, and Michael~S Bernstein.
\newblock Generative agents: Interactive simulacra of human behavior.
\newblock {\em arXiv preprint arXiv:2304.03442}, 2023.

\bibitem{xu2023exploring}
Yuzhuang Xu, Shuo Wang, Peng Li, Fuwen Luo, Xiaolong Wang, Weidong Liu, and Yang Liu.
\newblock Exploring large language models for communication games: An empirical study on werewolf.
\newblock {\em arXiv preprint arXiv:2309.04658}, 2023.

\bibitem{chen2023put}
Jiangjie Chen, Siyu Yuan, Rong Ye, Bodhisattwa~Prasad Majumder, and Kyle Richardson.
\newblock Put your money where your mouth is: Evaluating strategic planning and execution of llm agents in an auction arena.
\newblock {\em arXiv preprint arXiv:2310.05746}, 2023.

\bibitem{ge2023openagi}
Yingqiang Ge, Wenyue Hua, Kai Mei, Jianchao Ji, Juntao Tan, Shuyuan Xu, Zelong Li, and Yongfeng Zhang.
\newblock Open{AGI}: When {LLM} meets domain experts.
\newblock In {\em Thirty-seventh Conference on Neural Information Processing Systems}, 2023.

\bibitem{du2023improving}
Yilun Du, Shuang Li, Antonio Torralba, Joshua~B. Tenenbaum, and Igor Mordatch.
\newblock Improving factuality and reasoning in language models through multiagent debate, 2023.

\bibitem{chan2023chateval}
Chi-Min Chan, Weize Chen, Yusheng Su, Jianxuan Yu, Wei Xue, Shanghang Zhang, Jie Fu, and Zhiyuan Liu.
\newblock Chateval: Towards better llm-based evaluators through multi-agent debate, 2023.

\bibitem{sun2023corex}
Qiushi Sun, Zhangyue Yin, Xiang Li, Zhiyong Wu, Xipeng Qiu, and Lingpeng Kong.
\newblock Corex: Pushing the boundaries of complex reasoning through multi-model collaboration, 2023.

\bibitem{liang2023encouraging}
Tian Liang, Zhiwei He, Wenxiang Jiao, Xing Wang, Yan Wang, Rui Wang, Yujiu Yang, Zhaopeng Tu, and Shuming Shi.
\newblock Encouraging divergent thinking in large language models through multi-agent debate, 2023.

\bibitem{wang2023humanoid}
Zhilin Wang, Yu~Ying Chiu, and Yu~Cheung Chiu.
\newblock Humanoid agents: Platform for simulating human-like generative agents, 2023.

\bibitem{fu2023improving}
Yao Fu, Hao Peng, Tushar Khot, and Mirella Lapata.
\newblock Improving language model negotiation with self-play and in-context learning from ai feedback.
\newblock {\em arXiv preprint arXiv:2305.10142}, 2023.

\bibitem{hong2023metagpt}
Sirui Hong, Xiawu Zheng, Jonathan Chen, Yuheng Cheng, Ceyao Zhang, Zili Wang, Steven Ka~Shing Yau, Zijuan Lin, Liyang Zhou, Chenyu Ran, et~al.
\newblock Metagpt: Meta programming for multi-agent collaborative framework.
\newblock {\em arXiv preprint arXiv:2308.00352}, 2023.

\bibitem{liu2023bolaa}
Zhiwei Liu, Weiran Yao, Jianguo Zhang, Le~Xue, Shelby Heinecke, Rithesh Murthy, Yihao Feng, Zeyuan Chen, Juan~Carlos Niebles, Devansh Arpit, et~al.
\newblock Bolaa: Benchmarking and orchestrating llm-augmented autonomous agents.
\newblock {\em arXiv preprint arXiv:2308.05960}, 2023.

\bibitem{qian2023communicative}
Chen Qian, Xin Cong, Wei Liu, Cheng Yang, Weize Chen, Yusheng Su, Yufan Dang, Jiahao Li, Juyuan Xu, Dahai Li, Zhiyuan Liu, and Maosong Sun.
\newblock Communicative agents for software development, 2023.

\bibitem{babyagi2023}
Yohei Nakajima.
\newblock Babyagi.
\newblock \url{https://github.com/yoheinakajima/babyagi}, 2023.

\bibitem{chen2023agentverse}
Weize Chen, Yusheng Su, Jingwei Zuo, Cheng Yang, Chenfei Yuan, Chen Qian, Chi-Min Chan, Yujia Qin, Yaxi Lu, Ruobing Xie, et~al.
\newblock Agentverse: Facilitating multi-agent collaboration and exploring emergent behaviors in agents.
\newblock {\em arXiv preprint arXiv:2308.10848}, 2023.

\bibitem{li2023camel}
Guohao Li, Hasan Abed Al~Kader Hammoud, Hani Itani, Dmitrii Khizbullin, and Bernard Ghanem.
\newblock Camel: Communicative agents for "mind" exploration of large scale language model society, 2023.

\bibitem{dickson2002road}
Ted Dickson.
\newblock The road to united states involvement in world war i: A simulation.
\newblock {\em OAH Magazine of History}, 17(1):48--56, 2002.

\bibitem{guetzkow1963simulation}
Harold~Steere Guetzkow, Chadwick~F Alger, and Richard~A Brody.
\newblock Simulation in international relations: Developments for research and teaching.
\newblock {\em (No Title)}, 1963.

\bibitem{tollefson2008onesaf}
Eric Tollefson, M~Martin, Andrew Fletcher, and ARMY TRADOC ANALYSIS CENTER~MONTEREY CA.
\newblock Onesaf objective system (oos) behavior model verification.
\newblock {\em US Army TRADOC Analysis Center--Monterey, Monterey, CA}, 2008.

\bibitem{kellycreating}
Kevin~M Kelly, Craig Finch, David Tartaro, and Shiva Jaganathan.
\newblock Creating a world war ii combat simulator using onesaf objective system.
\newblock In {\em Proceedings of the Interservice/Industry Training, Simulation, and Education Conference}, pages 510--520, 2006.

\bibitem{hill2004using}
Raymond~R Hill, Lance~E Champagne, and Joseph~C Price.
\newblock Using agent-based simulation and game theory to examine the wwii bay of biscay u-boat campaign.
\newblock {\em The Journal of Defense Modeling and Simulation}, 1(2):99--109, 2004.

\bibitem{liu2023lost}
Nelson~F Liu, Kevin Lin, John Hewitt, Ashwin Paranjape, Michele Bevilacqua, Fabio Petroni, and Percy Liang.
\newblock Lost in the middle: How language models use long contexts.
\newblock {\em arXiv preprint arXiv:2307.03172}, 2023.

\bibitem{yang2023can}
Chenghao Yang and Allyson Ettinger.
\newblock Can you follow me? testing situational understanding in chatgpt.
\newblock {\em arXiv preprint arXiv:2310.16135}, 2023.

\bibitem{mombauer2013origins}
Annika Mombauer.
\newblock {\em The origins of the First World War: controversies and consensus}.
\newblock Routledge, 2013.

\bibitem{pick1960pulled}
Otto Pick.
\newblock Who pulled the trigger: Soviet historians and the origins of world war ii.
\newblock {\em Probs. Communism}, 9:64, 1960.

\bibitem{juliano1991warring}
Annette~L Juliano.
\newblock The warring states period—the state of qin, yan, chu, and pazyryk: A historical footnote.
\newblock {\em Source: Notes in the History of Art}, 10(4):25--29, 1991.

\bibitem{brown2020language}
Tom Brown, Benjamin Mann, Nick Ryder, Melanie Subbiah, Jared~D Kaplan, Prafulla Dhariwal, Arvind Neelakantan, Pranav Shyam, Girish Sastry, Amanda Askell, et~al.
\newblock Language models are few-shot learners.
\newblock {\em Advances in neural information processing systems}, 33:1877--1901, 2020.

\bibitem{openai2023gpt4}
OpenAI.
\newblock Gpt-4 technical report, 2023.

\bibitem{claude}
Anthropic.
\newblock Model card and evaluations for claude models.
\newblock 2023.

\bibitem{keegan2014first}
John Keegan.
\newblock {\em The first world war}.
\newblock Random House, 2014.

\bibitem{vinh2010bailey}
Epps Vinh.
\newblock Bailey,(2010).
\newblock {\em Information Theoretic Measures for Clusterings Comparison: Variants, Properties, Normalization and Correction for Chance. JMLR}, pages 2837--2854, 2010.

\bibitem{rajaraman2011mining}
Anand Rajaraman and Jeffrey~David Ullman.
\newblock {\em Mining of massive datasets}.
\newblock Cambridge University Press, 2011.

\bibitem{epstein2023inverse}
Joshua~M Epstein.
\newblock Inverse generative social science: Backward to the future.
\newblock {\em Journal of artificial societies and social simulation: JASSS}, 26(2), 2023.

\bibitem{dal2002multi}
Arianna Dal~Forno, Ugo Merlone, et~al.
\newblock A multi-agent simulation platform for modeling perfectly rational and bounded-rational agents in organizations.
\newblock {\em JASSS}, 5:1--16, 2002.

\bibitem{polhill2007open}
J~Gary Polhill and BM~Edmonds.
\newblock Open access for social simulation.
\newblock {\em Journal of Artificial Societies and Social Simulation}, 10(3):10--10, 2007.

\bibitem{national2006defense}
National~Research Council et~al.
\newblock {\em Defense modeling, simulation, and analysis: meeting the challenge}.
\newblock National Academies Press, 2006.

\bibitem{stauffer2002sociophysics}
Dietrich Stauffer.
\newblock Sociophysics simulations.
\newblock {\em arXiv preprint cond-mat/0210213}, 2002.

\bibitem{sun2006cognition}
Ron Sun.
\newblock {\em Cognition and multi-agent interaction: From cognitive modeling to social simulation}.
\newblock Cambridge University Press, 2006.

\end{thebibliography}
\end{document}